\documentclass[10pt,twocolumn,letterpaper]{article}

\usepackage{wacv}
\usepackage[accsupp]{axessibility}
\usepackage{times}
\usepackage{epsfig}
\usepackage{graphicx}
\usepackage{amsmath}
\usepackage{amssymb}

\usepackage{float}
\usepackage{graphicx}
\usepackage{color}
\usepackage{xcolor}
\usepackage{bm}
\usepackage{times}
\usepackage{epsfig}
\newcommand{\tabincell}[2]{\begin{tabular}{@{}#1@{}}#2\end{tabular}}
 \usepackage{transparent}
\usepackage{mathrsfs}
\usepackage{dashrule}

\usepackage{multirow}
\usepackage{bbding}
\usepackage[flushleft]{threeparttable}
\usepackage{caption}
\usepackage{subcaption}
\usepackage{tabularx}
\usepackage{placeins}
\usepackage{stfloats}

\usepackage{booktabs}

\makeatletter
\newcommand{\thickhline}{%
    \noalign {\ifnum 0=`}\fi \hrule height 1pt
    \futurelet \reserved@a \@xhline
}

\def\wacvPaperID{492} 

\wacvfinalcopy 

\ifwacvfinal
\def\assignedStartPage{9876} 
\fi

\usepackage[pagebackref=true,breaklinks=true,colorlinks,bookmarks=false]{hyperref}
\ifwacvfinal
\else
\fi

\begin{document}

\title{Busy-Quiet Video Disentangling for Video Classification}

\author{Guoxi Huang and Adrian G. Bors\\
\vspace*{-0.2cm}\\
Department of Computer Science\\
University of York, York YO10 5GH, UK\\
{\tt\small \{gh825, adrian.bors\}@york.ac.uk}
}

\maketitle
\thispagestyle{empty}

\begin{abstract}
In video data, busy motion details from moving regions are conveyed within a specific frequency bandwidth in the frequency domain. Meanwhile, the rest of the frequencies of video data are encoded with quiet information with substantial redundancy, which causes low processing efficiency in existing video models that take as input raw RGB frames. In this paper, we consider allocating intenser computation for the processing of the important busy information and less computation for that of the quiet information. We design a trainable Motion Band-Pass Module (MBPM) for separating busy information from quiet information in raw video data.
By embedding the MBPM into a two-pathway CNN architecture, we define a Busy-Quiet Net (BQN). The efficiency of BQN is determined by avoiding redundancy in the feature space processed by the two pathways: one operating on Quiet features of low-resolution, while the other processes Busy features.
The proposed BQN outperforms many recent video processing models on Something-Something V1, Kinetics400, UCF101 and HMDB51 datasets. The code is available at: \href{https://github.com/guoxih/busy-quiet-net}{\textit{https://github.com/guoxih/busy-quiet-net}}.
\end{abstract}

\section{Introduction}
\label{sec:intro}
Video classification is a fundamental problem in many video-based tasks.
Applications such as autonomous driving technology, controlling drones and robots are driving the demand for new video processing methods. 
An effective way to extend the usage of Convolutional Neural Networks (CNNs) from the image to the video domain is by expanding the convolution kernels from 2D to 3D~\cite{carreira2017quo,hara2018can,taylor2010convolutional,tran2015learning}.
Since the progress made by I3D~\cite{carreira2017quo}, the main research effort in the video area has been directed towards designing new 3D architectures.
However, 3D CNNs are more computationally intensive than 2D CNNs. Some recent works \cite{feichtenhofer2020x3d,lin2019tsm,qiu2017learning,tran2019video,tran2018closer,xie2018rethinking} increase the efficiency of 3D CNNs by reducing the redundancy in the model parameters. However, these works have ignored another important factor that causes the heavy computation in video processing: natural video data contains substantial redundancy in the spatio-temporal dimensions.
\begin{figure}[!t]
  \centering
    \resizebox{.4\textwidth}{!}{\includegraphics{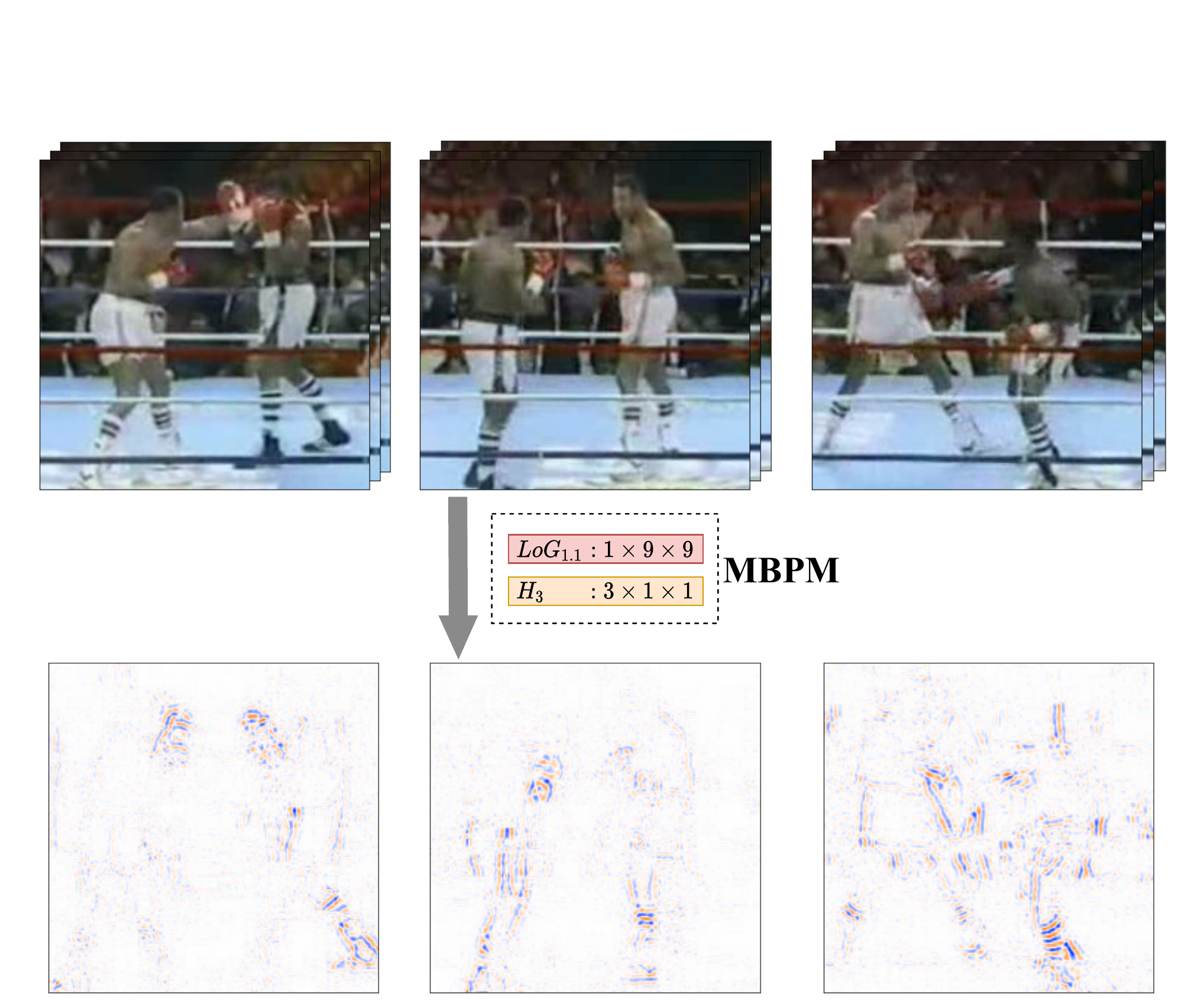}}
  \caption{Motion Band-Pass Module (MBPM) distills busy information from the frame sequences. For every three consecutive RGB frames, the MBPM generates a one-frame output, which substantially reduces the redundancy.}
\label{fig:mbpm}
\vspace{-0.35cm}
\end{figure}
In video data, busy information describes fast-changing motion happening in the boundaries of moving regions which is crucial for defining movement in video. Meanwhile, the quiet information, such as smooth background textures whose information is shared by neighboring locations, contains substantial redundancy. 
For efficient processing, we disentangle a video into busy and quiet components. 
Subsequently, we would separately process the busy and quiet components, by allocating high-complexity processing for the busy information and low-complexity processing for the quiet information. 

\begin{figure*}[!t]
  \centering
    \includegraphics[width=17cm,height=6.2cm]{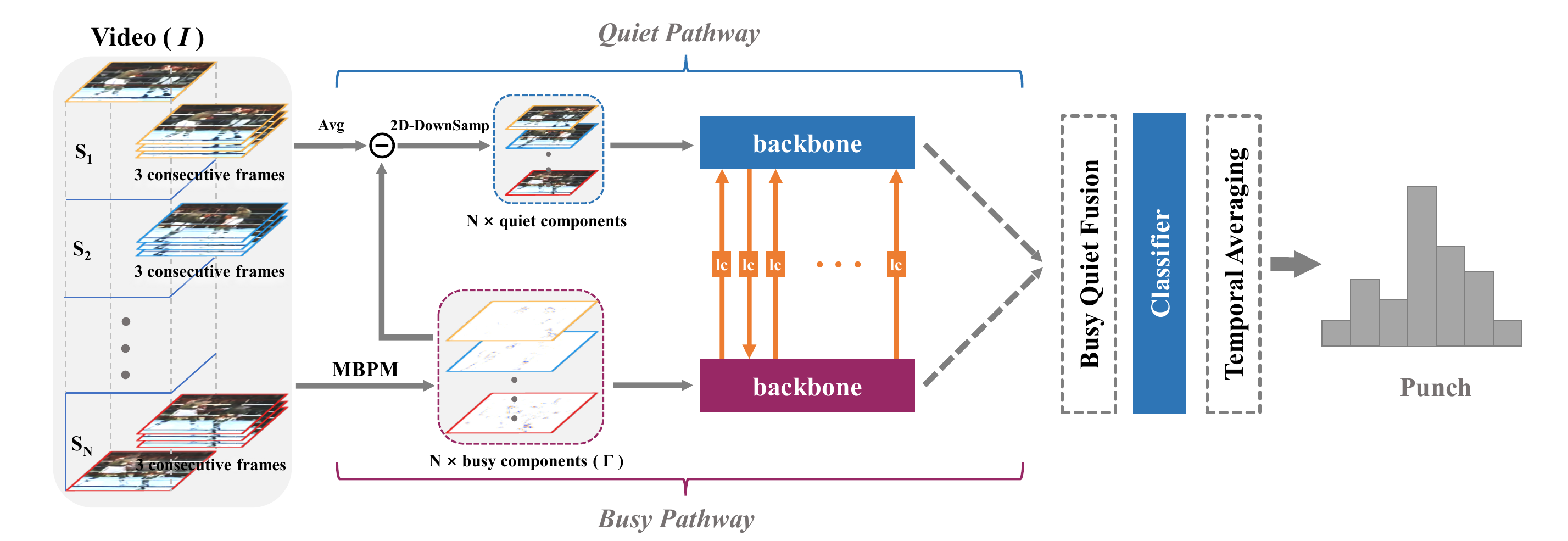}
  \caption{BQN is made up of two parallel pathways: Busy and Quiet. `lc' indicates Band-Pass Lateral Connection. The backbone networks from the two pathways respectively take as inputs two complementary data components (\ie busy and quiet), which are disentangled by the MBPM. The outputs of the two pathways are fused at various processing stages and the final prediction is obtained by averaging the prediction scores across multiple segments.}
\label{fig:framework}
\end{figure*}

In this study we propose a lightweight, end-to-end trainable motion feature extraction mechanism called Motion Band-Pass Module (MBPM), which can distill the motion information conveyed within a specific frequency bandwidth in the frequency domain. As illustrated in  Figure~\ref{fig:mbpm}, by applying the MBPM to a video of 3 segments the number of representative frames is reduced from 9 to 3 whilst retaining the essential motion information.
Our experiments demonstrate that by simply replacing the RGB frame input with the motion representation extracted by our MBPM, the performance of existing video models can be boosted. 
Secondly, we design a two-pathway multi-scale architecture, called the Busy-Quiet Net (BQN), the processing pipeline of which is shown in Figure~\ref{fig:framework}. One pathway, called Busy, is responsible for processing the busy information distilled by the MBPM. The other pathway, called Quiet, is devised to process the quiet information encoded with global smooth spatio-temporal structures. In order to fuse the information from different pathways, we design the Band-Pass Lateral Connection (BPLC) module, which is set up between the Busy and Quiet pathways. During the experiments, we demonstrate that the BPLC is the key factor to the overall model optimization success. 

Compared with the frame summarization approaches~\cite{bilen2017action,wang2018video},  MBPM retains the strict temporal order of the frame sequences, which is considered essential for long-term temporal relation modeling.
Compared with optical flow-based motion representation methods~\cite{fan2018end,ilg2017flownet,wang2016temporal,zach2007duality,zhang2019pan}, the motion representation captured by MBPM has a smaller temporal size (\eg for every 3 RGB frames, the MBPM encodes only one frame), and can be employed on the fly. Meanwhile, efficient video models such as Octave Convolution~\cite{chen2019drop}, bL-Net~\cite{chen2018big} and SlowFast networks~\cite{feichtenhofer2019slowfast} only reduce the input redundancy along either the spatial or temporal dimensions. Instead, the proposed BQN reduces the redundancy in the joint spatio-temporal space. 

Our contributions can be summarized as follows:
\begin{itemize}
\vspace*{-0.23cm}
\item A novel Motion Band-Pass Module (MBPM) is proposed for busy motion information distillation. The new motion cue extracted by the MBPM is shown to reduce temporal redundancy significantly.

\item \vspace*{-0.23cm} We design a two-pathway Busy-Quiet Net (BQN) that separately processes the busy and quiet information in videos. By separating the busy information using MBPM, we can safely downsample the quiet information to further reduce redundancy.

\item  \vspace*{-0.23cm} Extensive experiments demonstrate the superiority of the proposed BQN over a wide range of models on four standard video benchmarks: Kinetics400~\cite{carreira2017quo}, Something-Something V1~\cite{goyal2017something}, UCF101~\cite{soomro2012ucf101}, HMDB51~\cite{kuehne2011hmdb}.
\end{itemize}

\section{Related Work}
\label{Related Work}

\noindent \textbf{Spatio-temporal Networks.} 
Early works~\cite{donahue2015long,feichtenhofer2016convolutional,simonyan2014two,wang2016temporal,yue2015beyond} attempt to extend the success of 2D CNNs~\cite{he2016deep,huang2017densely,krizhevsky2012imagenet,simonyan2014very,szegedy2015going,tan2019efficientnet} in the image domain to the video domain. Representatively, the two-stream model~\cite{simonyan2014two} and its variants~\cite{feichtenhofer2016convolutional,wang2016temporal} utilize optical flow as an auxiliary input modality for effective temporal modeling.
Other works~\cite{carreira2017quo,hara2018can,taylor2010convolutional,tran2015learning}, given the progress in GPU performance, tend to exploit the computationally intensive 3D convolution. Meanwhile, some studies focus on improving the efficiency of 3D CNN, such as P3D~\cite{qiu2017learning}, R(2+1)D~\cite{tran2018closer}, S3D~\cite{xie2018rethinking}, TSM~\cite{lin2019tsm}, CSN~\cite{tran2019video},  X3D~\cite{feichtenhofer2020x3d}. Non-local Net~\cite{wang2018non} and its variants~\cite{cao2019gcnet,huang2021icpr} introduce self-attention mechanisms to CNNs in order to learn long-range dependencies in the spatio-temporal dimension. Our study is complementary to these methods: our Busy-Quiet Net (BQN) can benefit from the efficiency of these CNNs by simply adopting them as backbones.

\noindent \textbf{Motion Representation.}
Optical flow as a short-term motion representation has been widely used in many video applications. However, the optical flow estimation in large-scale video datasets is inefficient. Some recent works use deep learning to improve the optical flow estimation quality, such as FlowNet  ~\cite{dosovitskiy2015flownet,ilg2017flownet}.
Some other methods aim to explore new end-to-end trainable motion cues, such as OFF~\cite{sun2018optical}, TVNet~\cite{fan2018end}, EMV~\cite{zhang2018real}, Flow-of-Flow~\cite{piergiovanni2019representation}, Dynamic Image~\cite{bilen2017action}, Squeezed Image~\cite{huang2020learning} and PA~\cite{zhang2019pan}. Compared with these approaches, our MBPM is rather as a basic video architecture component which results in higher accuracy while requiring less computation.

\noindent \textbf{Enforcing Low Information Redundancy.}
In the image field, bL-Net~\cite{chen2018big} adopts a downsampling strategy that operates at the block level aiming to reduce the spatial redundancy of its feature maps.
Octave Convolution~\cite{chen2019drop} replaces the convolutions in existing CNNs to decompose the low and high-frequency components in images, representing the former with lower resolution. In the video field, bLV-Net~\cite{fan2019more} extends the idea of bL-Net~\cite{chen2018big} to the temporal dimension. SlowFast networks~\cite{feichtenhofer2019slowfast} introduce two pathways for slow and fast motion decomposition along the temporal dimension. However, the generalization of SlowFast to existing CNN architectures is poor, as it requires two specially customized CNNs to be its backbones. 
Unlike the previous methods, BQN reduces the feature redundancy in the joint spatio-temporal space by using a predefined trainable filter module, MBPM, to disentangle a video into busy and quiet components. Unlike SlowFast, BQN architecture shows excellent generalization to existing CNNs.

\section{Motion Band-Pass Module (MBPM)}
\label{sec:mbpm}
Firstly, we introduce a 3D band-pass filter, which can distill the video motion information conveyed within a specific spatio-temporal frequency bandwidth.
A video clip of $T$ frames can be defined as a function with three arguments, $\boldsymbol{I}^{(t)}(x, y)$, where $x$, $y$ indicate the spatial dimensions, while $t$ is the temporal dimension. The value of $\boldsymbol{I}^{(t)}(x, y)$ corresponds to the pixel value at position $(x,y)$ in $t$-th frame of an arbitrary channel in the video.  When considering the multi-channel case, we repeat the same procedure for each color channel, which is omitted here for the sake of simplification.
The output $\boldsymbol{\Gamma}$ of the 3D band-pass filter is given by:
\begin{equation}
\begin{aligned}
\boldsymbol{\Gamma}(x,y,t) &= \frac{\partial^2}{\partial t^2} \left[ \boldsymbol{I}^{(t)}(x,y) * LoG_\sigma(x,y) \right], \\
&\approx \sum\limits_{t-1\leq i\leq t+1} h{\scriptstyle(i)} \cdot [\boldsymbol{I}^{(i)}(x,y) * LoG_\sigma(x,y)],\\
& \qquad \qquad \qquad \qquad \qquad h{\scriptstyle(i)} =
\begin{cases}
 \frac{2}{3} & \text{if} \quad i = t, \\
-\frac{1}{3} & \text{otherwise},
\end{cases}
\end{aligned}
\label{eq: mbpm}
\end{equation}
for $t= 1,\ldots,T$ and `$*$' represents the convolution operation. 
In equation~\eqref{eq: mbpm}, the second derivative with respect to $t$ is numerically approximated by finite differences, literally implemented by function $h{\scriptstyle(i)}$.
Meanwhile, $LoG_\sigma(x,y)$ is a two-dimensional Laplacian of Gaussian with the scale parameter $\sigma$:
\begin{equation}
\begin{aligned}
LoG_\sigma(x,y) &= \triangledown^2G_\sigma(x,y)  
= -\frac{e^{-\frac{x^2+y^2}{2\sigma^2}}}{\pi \sigma^4} \left[ 1- \frac{x^2+y^2}{2\sigma^2} \right]. 
\end{aligned}
\label{eq:log}
\end{equation}

From equations \eqref{eq: mbpm} and \eqref{eq:log} we can observe that the 3D filtering function is fully-differentiable. In order to make the 3D band-pass filtering compatible with CNNs, we approximate it with two sequential channel-wise\footnote{Also referred to as “depth-wise". We use the term “channel-wise" to avoid confusions with the network depth.} convolutional layers~\cite{sandler2018mobilenetv2}, as shown in Figure~\ref{fig:mbpm}. We name the discrete approximation Motion Band-Pass Module (MBPM) which can be expressed in an engineering form as follows:
\begin{equation}
\begin{aligned}
\boldsymbol{\Gamma} \approx \textup{MBPM}(\boldsymbol{I}) = H_{s\times 1 \times 1}^{3\times 1 \times 1}(LoG_\sigma^{1\times k \times k}(\boldsymbol{I})),
\end{aligned}
\label{eq:mbpm_engi}
\end{equation}
where $LoG_\sigma^{1\times k \times k}$ is referred to as a spatial channel-wise convolutional layer~\cite{sandler2018mobilenetv2} with a $k \times k$ kernel, each channel of which is initialized with a Laplacian of Gaussian distribution with scale $\sigma$. The sum of kernel values is normalized to 1. Meanwhile, $H_{s\times 1 \times 1}^{3\times 1 \times 1}$ is referred to as a temporal channel-wise convolutional layer with a temporal stride $s$. In each channel, the kernel value of $H_{s\times 1 \times 1}^{3\times 1 \times 1}$ is initialized with $[-\frac{1}{3}, \frac{2}{3}, -\frac{1}{3}]$ , which is a high-pass filter.
In order to let the MBPM kernel parameters fine-tune on video streams, we embed the MBPM within a CNN for end-to-end training, optimized with the video classification loss.

\section{Busy-Quiet Net (BQN)}
\label{sec:BQN}

 As illustrated in Figure~\ref{fig:framework}, the BQN architecture contains two pathways, Busy and Quiet, operating in parallel on two distinct video data components, which are separated by the MBPM.
 The Busy and Quiet pathways are bridged by multiple Band-Pass Lateral Connections (see Section~\ref{sec:lc}). These lateral connections enable information fusion between the two processing pathways.

\subsection{Busy and Quiet pathways}
\label{sec:BusyandQuiet}

\noindent \textbf{Busy pathway.} 
The Busy pathway is designed to learn fine-grained movement features. It takes as input the information filtered by the MBPM, which contains critical motion information located at the boundaries of objects or regions that have significant temporal change.
The stride of $H_{s\times 1 \times 1}^{3\times 1 \times 1}$ from equation~\eqref{eq:mbpm_engi} is set to $s=3$, which means that for every three consecutive RGB frames, the MBPM generates one-frame output. The MBPM output preserves the temporal order of the video frames while significantly reducing the redundant temporal information.
We intend to utilize larger spatial input sizes for the Busy pathway to extract more distinct textures.

\noindent \textbf{Quiet pathway.} The Quiet pathway focuses on processing quiet information, representing the characteristics of large regions of movement, such as the movement happening in the plain-textured background regions. The input to the Quiet pathway is considered to be the complementary of the MBPM output:
\begin{equation}
\begin{aligned}
\textup{2D-DownSamp}(\textup{Avg}_{3\times 1 \times 1}^{3\times 1 \times 1} (\boldsymbol{I}) - \boldsymbol{\Gamma}),
\end{aligned}
\label{eq:quiet_input}
\end{equation}
where $\textup{Avg}_{3\times 1 \times 1}^{3\times 1 \times 1}$ is a temporal average pooling with a stride of 3. In the spatial dimensions, we perform bilinear downsampling (\ie $\textup{2D-DownSamp}$) to reduce the redundant spatial information shared by neighboring locations. In Section~\ref{sec:ablation_BQN}, we explore the effect on performance when varying the input size of the Quiet pathway.

\subsection{Band-Pass Lateral Connection (BPLC)}
\label{sec:lc}
 \vspace*{-0.2cm}
In the proposed BQN, we include a novel Band-Pass Lateral Connection (BPLC) module which has an MBPM embedded. The BPLCs established between the two pathways, Busy and Quiet, provide a mechanism for information exchange, enabling an optimal fusion of video information characterized by different frequency bands. Different from the lateral connections in~\cite{feichtenhofer2019slowfast,feichtenhofer2016spatiotemporal,feichtenhofer2016convolutional,lin2017feature}, the BPLC, enabled by MBPM, performs feature fusion and feature selection simultaneously, which shows higher performance than other lateral connection designs, according to the experimental results.
We denote the two inputs of BPLC from the $i$-th residual blocks in the Busy and Quiet pathways, as $\mathbf{x}_f^i$ and $\mathbf{x}_c^i$, respectively. For simplifying the notation, we assume that $\mathbf{x}_f^i$ and $\mathbf{x}_c^i$ are of the same size. When their sizes are different, we adopt bilinear interpolation to match them in size. The output $\mathbf{y}_f^i$ and $\mathbf{y}_c^i$ for the Busy and Quiet, respectively, is given by
\begin{equation}
\vspace*{-0.2cm}
\begin{aligned}
\mathbf{y}_f^i &= 
\begin{cases}
\mathrm{BN}(\textup{MBPM}(\mathbf{x}_c^i)) + \mathbf{x}_f^i  &  \text{if} \quad \text{mod}(i,2) = 0, \\
 \mathbf{x}_f^i & \text{otherwise},
\end{cases} \\
\mathbf{y}_c^i &=
\begin{cases}
 \mathbf{x}_c^i & \quad \,\,\,\,\,\,\,\,\,\,\text{if} \quad \text{mod}(i,2) = 0,\\
 \mathrm{BN}(\mathbf{\phi}(\mathbf{x}_f^i)) + \mathbf{x}_c^i  &  \quad \,\,\,\,\,\,\,\,\,\, \text{otherwise},
\end{cases} \\
& \quad \quad \quad \quad \quad \quad \quad \quad \quad \quad \quad \quad \,\,\,\,\, i = 1,2,\ldots,B
\end{aligned}
\label{eq:bplc}
\vspace*{+0.1cm}
 \end{equation}
where $B$ denotes the number of residual blocks in the backbone network (considered as the network with residual block designs in the experiments). $\mathbf{\phi}(\cdot)$ is the linear transformation that can be implemented as a $1\times1 \times 1$ convolution, or alternatively, when the channel number is very large, as a bottleneck MLP for reducing computation.  $\mathrm{BN}$ indicates Batch Normalization~\cite{szegedy2015bn}, the weights of which are initialized to zero. For the MBPM in BPLC, the convolutional stride of $H_{s \times 1 \times 1}^{3\times 1 \times 1}$ from equation~\eqref{eq:mbpm_engi} is set to $s=1$, maintaining the same temporal size.

The fusion direction of BPLC reverses back and forth, as indicated in Figure~\ref{fig:framework}, providing better communication for the two pathways than the unidirectional lateral connections in~\cite{feichtenhofer2019slowfast,lin2017feature} whose information fusion direction is fixed always fusing the information from a certain pathway to the other.
By default, we place a BPLC between the two pathways right after each pair of residual blocks. 
The MBPM embedded in BPLC acts as a soft feature selection gate, which allows only for the busy information from the Quiet pathway to flow to the Busy pathway during the information fusion process. 
The exploration of various lateral connection designs is provided in Section~\ref{sec:ablation_BQN} and this setting is shown to give the best performance in our experiments.
\begin{figure}[t]
    \centering
    \includegraphics[width=.93\linewidth]{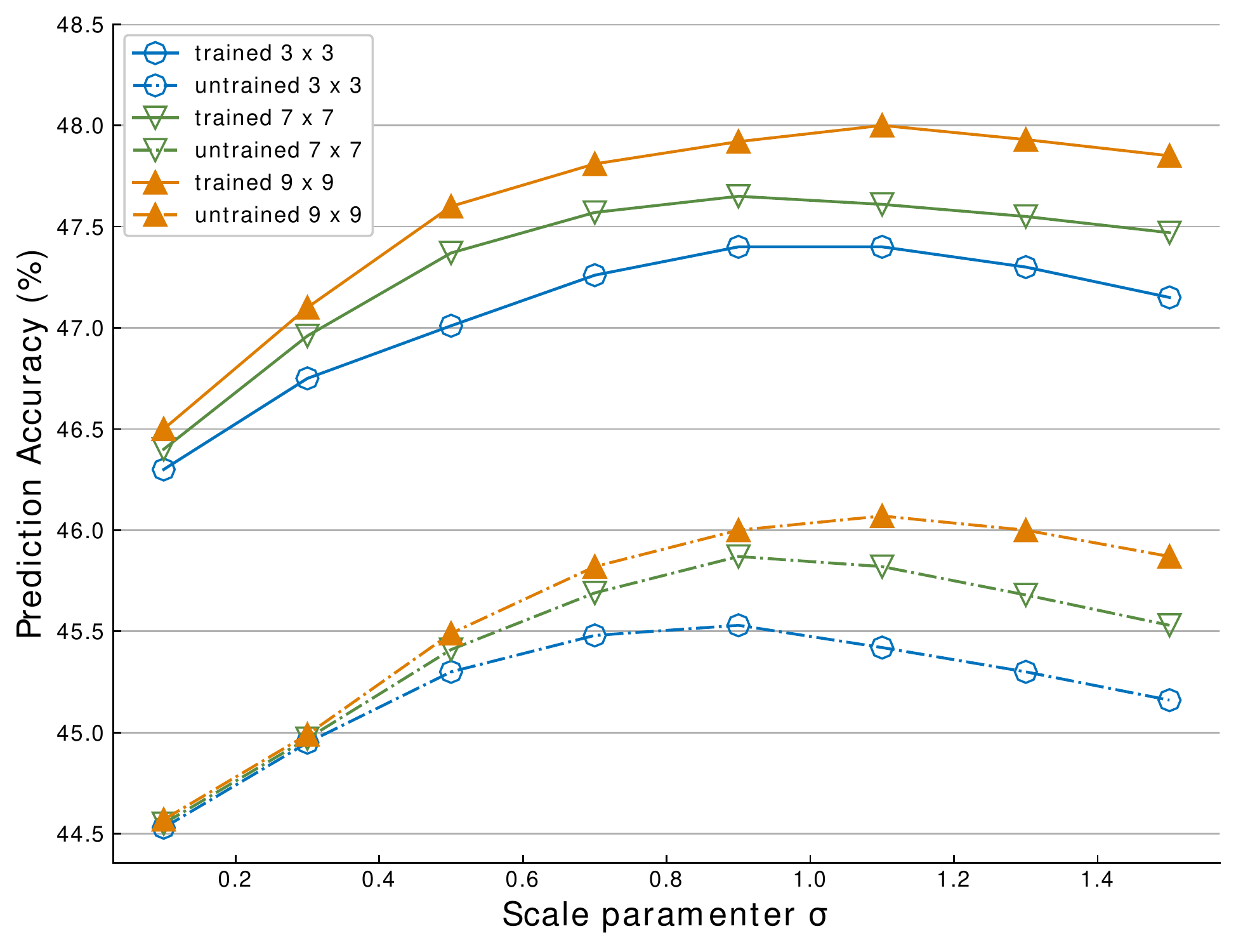}
    \vspace{-6pt}
    \caption{Results on SS V1 when varying the  scale $\sigma$ and kernel size $k \times k$ of the spatial channel-wise convolution in MBPM. The results represent averages of multiple runs.}
    \label{fig:sima_eva}
    \vspace{-10pt}
\end{figure}

\section{Experiments}

In this section, we provide the results of the experiments for the proposed MBPM and BQN. Then, we compare these results with the state-of-the-art.  Unless otherwise stated, we use ResNet50 (R50) with TSM ~\cite{lin2019tsm}, as the backbone of our model.
\begin{table}[t]
\small
\caption{MBPM \vs other motion representation methods.  $\dagger$ denotes our reimplementation. The additional parameters and the required computation (FLOPs) are reported.}
\vspace{-0.3cm}
\label{table:compar_mbpm}
\begin{center}
\setlength{\tabcolsep}{0pt}
\begin{tabular}{l c c c c c}
\toprule
\multirow{2}{2cm}{\bfseries Rep. Method} & \multicolumn{2}{c}{ \bfseries Efficiency Metrics}  & \multirow{2}{1.3cm}{ \bfseries UCF101} & \multirow{2}{1cm}{ \bfseries SS V1} & \multirow{2}{1cm}{ \bfseries K400}\\
\cmidrule{2-3}
 & \bfseries FLOPs & \bfseries \#Param. & & &\\
\midrule
RGB (baseline) &- &- & 87.1 & 46.5 & 71.2\\
RGB Diff~\cite{wang2016temporal}  &- & - &  87.0  & 46.6 & 71.4 \\
TV-L1 Flow~\cite{zach2007duality} &- & - & 88.5 & 37.4 & 55.7\\
\midrule
DI$\dagger$~\cite{bilen2017action} & - & - & 86.2 & 43.4& 68.3\\
FlowNetC$\dagger$~\cite{ilg2017flownet}  & 444G & 39.2M  & 87.3 & 26.3 & - \\
FlowNetS$\dagger$~\cite{ilg2017flownet}  & 356G &  38.7M & 86.8  & 23.4 & - \\
TVNet$\dagger$~\cite{fan2018end} & 3.3G & 0.2K & 88.6 & 45.2  & 58.5  \\
PA~\cite{zhang2019pan} & 2.8G&1.1K & 89.5 &  45.1 & 57.3 \\
\midrule
\textbf{MBPM} &0.3G & 0.2K & \textbf{90.3} & \textbf{48.0} & \textbf{72.3}\\
\bottomrule
\end{tabular}
\vspace{-8pt}
\end{center}
\end{table}
\begin{table}[!t]
\small
\caption{Using different CNNs as backbones on SS V1. ResNet50 and MobileNetV2 have TSM~\cite{lin2019tsm} embedded.}
\vspace{-0.3cm}
\label{tab:diff_cnns}
\begin{center}
\setlength{\tabcolsep}{2.5pt}
\begin{tabular}{l c c c c }
\toprule
\bfseries CNN backbone & \bfseries Modality  & \bfseries pretrain & \bfseries Seg. ($N$) & \bfseries Acc.  \\
\midrule
\multirow{4}*{ResNet50~\cite{he2016deep}} & RGB &  \multirow{4}*{ImageNet} & \multirow{4}*{8} & 46.5 \\
& RGB+Flow & & & 49.8 \\
& MBPM &  & & 48.0 \\
& RGB+MBPM &  & & 50.3 \\
\midrule
\multirow{2}*{MobileNetV2~\cite{sandler2018mobilenetv2} }& RGB &  \multirow{2}*{ImageNet} &  \multirow{2}*{8} &    38.7 \\
& MBPM &  & & 39.8 \\
\midrule
\multirow{2}*{X3D-M~\cite{feichtenhofer2020x3d} } & RGB & \multirow{2}*{None} &  \multirow{2}*{16} &  45.5 \\
& MBPM &  & & 46.9 \\
\bottomrule
\end{tabular}
\vspace{-0.4cm}
\end{center}
\end{table}

 \begin{figure*}[t]
\vspace*{-0.1cm}
    \centering
    \resizebox{1.\textwidth}{!}{\includegraphics{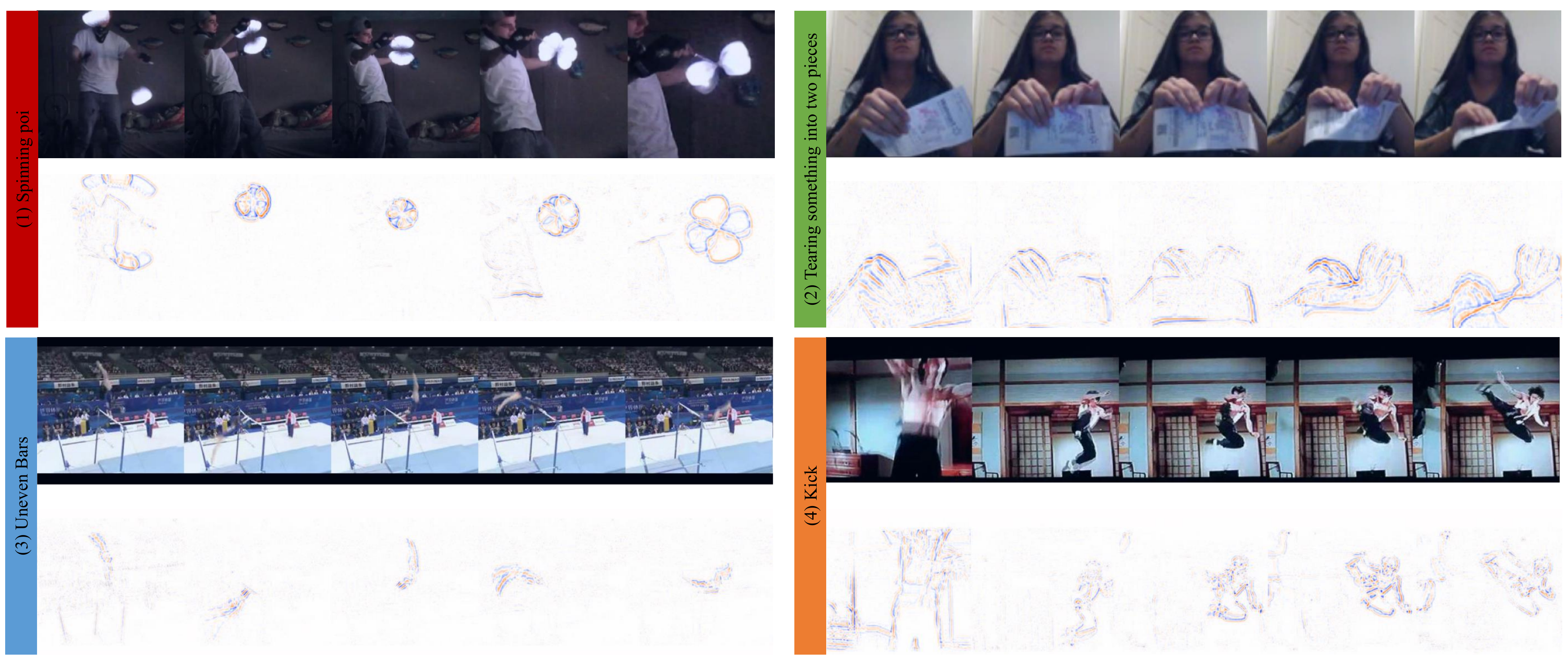}}
    \vspace*{-0.6cm}
    \caption{Videos and their MBPM outputs. The video clips (1)-(4) are from Kinetics, SS V1, UCF and HMDB, respectively. }
    \label{fig:example_mbpm}
    \vspace*{-0.4cm}
\end{figure*}

\subsection{Datasets and Implementation Details}
\noindent \textbf{Datasets}. We evaluate our approach on Something-Something V1~\cite{goyal2017something}, Kinetics400~\cite{carreira2017quo}, UCF101~\cite{soomro2012ucf101} and HMDB51~\cite{kuehne2011hmdb}.
Most of the videos in Kinetics400 (K400), UCF101 and HMDB51 can be accurately classified by only considering their background scene information, while the temporal relation between frames is not very important.
In Something-Something (SS) V1, many action categories are symmetrical ({\em e.g.} ``Pulling something from left to right'' and ``Pulling something from right to left''). Discriminating these symmetric actions requires models with strong temporal modeling ability. 

\noindent \textbf{Training \& Testing.}
Aside from X3D-M~\cite{feichtenhofer2020x3d}, the backbone networks are pretrained on ImageNet~\cite{imagenet_cvpr09}. For training, we utilize the dense sampling strategy~\cite{wang2018non} for 
Kinetics400. As for the other datasets, we utilize the uniform sampling strategy as shown in Figure~\ref{fig:framework}, where a video is equally divided into $N$ segments, and 3 consecutive frames in each segment are randomly sampled to constitute a video clip of length $T=3N$. 
Unless specified otherwise, a default video clip is composed of $N=8$ segments with a spatial size of $224^2$. 
During the tests, we sample a single clip per video with center cropping for efficient inference~\cite{lin2019tsm}, which is used in our ablation studies. When pursuing high accuracy, we consider sampling multiple clips\&crops from the video and averaging the prediction scores of multiple space-time ``views'' (spatial crops$\times$temporal clips) used in~\cite{feichtenhofer2019slowfast}. More training details can be found in Appendix~A.

\subsection{Ablation Studies for MBPM}
\label{sec: Ablation Studies for MBPM}

\noindent \textbf{Instantiations and Settings.} In the MBPM, the scale $\sigma$ from equation~\eqref{eq:log} and the kernel size of the spatial channel-wise convolution $LoG_\sigma^{1\times k \times k}$ have a significant impact on the performance. 
We vary the scale $\sigma $ and the kernel size to search for the optimal settings. Meanwhile, in order to highlight the importance of the training for the MBPM, we compare the performance when using trained MBPM with that of untrained MBPM whose kernel weights are not optimized with the classification loss. The results on SS V1 are shown in Figure~\ref{fig:sima_eva}. We summarize two facts: 1) the optimal value of $\sigma$ for the MBPM changes when the kernel size changes, and the MBPM with $\sigma=1.1$ and a spatial kernel size $9\times9$ gives the best performance within the searching range.  2) optimizing the parameters of MBPM with the video classification loss generally produces higher prediction accuracy.
In our preliminary work, we have verified that different datasets share the same optimal settings of MBPM.
More results on other datasets are provided in Appendix~B. 
In the following experiments, we set MBPM in the Busy pathway as trainable with the scale $\sigma=1.1$ and the kernel size of $9\times9$, unless specified otherwise.
\begin{table*}[t!]
\centering
\small
\captionsetup[subfloat]{labelformat=parens, labelsep=space, skip=7pt, position=bottom}
\caption{ Ablation Studies for BQN on Something-Something V1. We show top-1 and  top-5 prediction accuracy (\%), as well as computational complexity measured in GFLOPs for a single crop \& single clip.}
    \subfloat[\textbf{Complementarity of Quiet and Busy.}
    ``Quiet'' and ``Busy'' refer to that the Quiet and Busy pathways are trained separately. 
    \label{tab:abl_fcnet:fusion}]{
    \setlength{\tabcolsep}{3pt}
    \begin{tabularx}{0.31\linewidth}{l l c c c}
    \toprule
     & \bfseries Model &   \bfseries Top-1   &  \bfseries Top-5   & \bfseries GFLOPs\\
    \midrule
    &Quiet & 46.5 & 75.3 & 32.8\\
    &Busy & 48.0 & 76.8 & 32.8\\
    &Quiet+Busy& 50.3 & 79.0 & 65.7\\
    &BQN & \textbf{51.6} & \textbf{80.5} & 65.9\\
    \bottomrule
    \end{tabularx}}\hspace{2mm}
    \subfloat[\textbf{Effect of BPLCs.} The BQN model with more BPLCs has higher accuracy. \label{tab:abl_fcnet:num_lc}]{
    \setlength{\tabcolsep}{3pt}
    \begin{tabularx}{0.28\linewidth}{c c c c}
    \toprule
    \bfseries \# BPLC  & \bfseries \bfseries Top-1  & \bfseries \bfseries Top-5 & \bfseries GFLOPs\\
    \midrule
    0 & 49.6 & 78.9 & 65.7\\
    4 & 50.2 & 79.2 & 65.8\\
    8 & 50.7 & 79.7 & 65.8\\
    16 & \textbf{51.6} & \textbf{80.5} & 65.9\\
    \bottomrule
    \end{tabularx}}\hspace{2mm}
    \subfloat[\textbf{Fusion Strategies.}
    The fully-connected layers of the two pathways share the parameters. \label{tab:abl_fcnet:fusion_strategy}]{
    \setlength{\tabcolsep}{3pt}
    \begin{tabularx}{0.31\linewidth}{l c c c}
    \toprule
     \bfseries Fusion Method &  \bfseries Position & \bfseries \bfseries Top-1  &  \bfseries Top-5  \\
    \midrule
    Average & before fc  & 50.9 & 79.8  \\
    Average & after fc  & \textbf{51.6} & \textbf{80.5} \\
    Max & after fc  & 50.1 & 78.7 \\
    Concatenation & before fc & 51.3 & 80.2 \\
    \bottomrule
    \end{tabularx}}
    
    \subfloat[\textbf{Spatio-temporal input size.} The input size is formatted with ($\textup{width}^2 \times \textup{time}$). \label{tab:abl_fcnet: size}]{
    \setlength{\tabcolsep}{3pt}
    \begin{tabularx}{0.365\linewidth}{l c c c c}
    \toprule
    \tabincell{c}{\bfseries Input size   \\ \bfseries  for Quiet } &  \tabincell{c}{\bfseries Input size  \\ \bfseries for Busy } & \bfseries Top-1  & \bfseries Top-5  & \bfseries GFLOPs\\
    \midrule
    $224^2 \times 8$ & $224^2 \times 8$ & 51.6 & 80.5 & 65.9\\
    $160^2 \times 8$ & $224^2 \times 8$ & 51.3 & 80.1 & 50.5 \\
    $\mathbf{160^2\times8}$ & $\mathbf{256^2 \times 8}$ & \bfseries 51.7 & \bfseries 80.5 & \bfseries 60.7 \\
    $160^2 \times 6$ & $256^2 \times 8$ & 49.6 & 78.3 & 55.5 \\
    \bottomrule
    \end{tabularx}}\hspace{2mm}
    \subfloat[\textbf{Various LC designs.} 16 LCs are set in the BQN.
    \label{tab:abl_fcnet: lc_variants}]{
    \setlength{\tabcolsep}{5pt}
    \begin{tabularx}{0.21\linewidth}{l c c c c}
    \toprule
    \bfseries Design &  \bfseries Top-1   &  \bfseries Top-5 \\
    \midrule
    LC-I & 50.9 & 79.8 \\
    LC-II & 50.9 & 79.7 \\
    LC-III & 51.5 & 80.2 \\
    BPLC & \bfseries 51.6 & 80.5 \\
    LC-V & 51.3 & 79.9 \\
    \bottomrule
    \end{tabularx}}\hspace{2mm}
    \subfloat[\textbf{Stage for adding BPLCs.} In each stage, we set one BPLC after its first residual block.
    \label{tab:abl_fcnet: lc_stage}]{
    \setlength{\tabcolsep}{2pt}
    \begin{tabularx}{0.33\linewidth}{l c c c}
    \toprule
     \bfseries Stages & \bfseries \# BPLC & \bfseries Top-1 &  \bfseries Top-5\\
    \midrule
    res2  & 1 & 49.8 & 79.1 \\
    res2,res3  & 2 & 50.1 & 78.7 \\
    res2,res3,res4 & 3 & 50.2 & 79.0 \\
    res2,res3,res4,res5 & 4 & 50.2 & 79.2 \\
    \\
    \bottomrule
    \end{tabularx}}
    \label{tab: ablation_fcnet}
\end{table*}
\begin{figure*}[!t]
  \centering
    \resizebox{1.\textwidth}{!}{\includegraphics{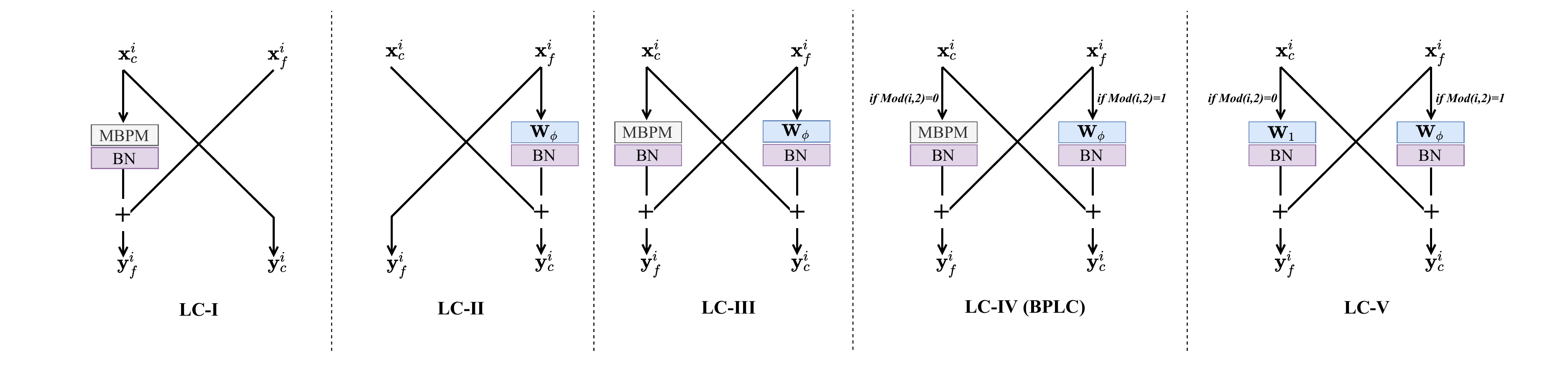}}
    \vspace{-0.5cm}
  \caption{Diagrams of various lateral connection (LC) designs. Bilinear interpolation is used for resizing the feature maps when $\mathbf{x}^i_c$ and $\mathbf{x}^i_f$ do not have the same spatial size. $i$ refers to the index of the residual block. $\mathbf{W}_\phi$ and $\mathbf{W}_1$ denote the weights of the linear transformation.}
  \vspace{-0.5cm}
\label{fig:lc_variant}
\end{figure*}

\noindent \textbf{Efficiency and Effectiveness of the MBPM.} 
We draw an apple-to-apple comparison between the proposed MBPM and other motion representation methods \cite{bilen2017action,fan2018end,ilg2017flownet,wang2016temporal,zach2007duality,zhang2019pan}. The comparison results are shown in Table~\ref{table:compar_mbpm}. The motion representations produced by these methods are used as inputs to the backbone network. The prediction scores are obtained by the average consensus of eight temporal segments~\cite{wang2016temporal}. More details about the implementations can be found in Appendix~C. The proposed MBPM outperforms the other motion representation methods by big margins, while its computation is nearly negligible, which strongly demonstrates the high efficiency and effectiveness of the MBPM. Moreover, the two-stream fusion of ``RGB+MBPM" has higher accuracy than the fusion of ``RGB+Flow", according to the results in Table~\ref{tab:diff_cnns}.

\noindent \textbf{Generalization to different CNNs.}
The proposed MBPM is a generic plug-and-play unit. The performance of existing video models could be boosted by simply placing an MBPM after their input layers. Table~\ref{tab:diff_cnns} show that MobileNetV2~\cite{sandler2018mobilenetv2} and X3D-M~\cite{feichtenhofer2020x3d} have steady performance improvement after being equipped with our MBPM.

\noindent \textbf{Visualization Analysis.}
We provide the visualization results of four videos and their corresponding MBPM outputs in the top and bottom rows from Figures~\ref{fig:example_mbpm} (1)-(4). From these results it can be observed that the extracted representations are stable when jittering and other camera movements are present. Also the results from Figures~\ref{fig:example_mbpm} (1)-(4) show that MBPM not only suppresses the stationary information and the background movement, but also highlights the boundaries of moving objects, which are of vital importance for action discrimination. For example, in the ``spinning poi'' video, from Figure~\ref{fig:example_mbpm}-(1), MBPM highlights the poi's movement rather than the movement of the background or that of the performer. More visualization results and visual comparison with other motion representation methods are provided in Appendix~D.

\subsection{Ablation Studies for BQN}
\label{sec:ablation_BQN}

\begin{table*}[!t]
\small
     \caption{\label{tab:sota_something} Results on Something V1. ``N/A'' indicates the numbers are not available. $\dagger$ denotes our reimplementation.
     }
     \vspace{-0.5cm}
      \begin{center}
      \setlength{\tabcolsep}{5pt}
      \begin{tabular}{l c c c c c c c}
      \toprule
      \bfseries Method & \bfseries Pretrain &\bfseries Backbone & \bfseries Frames$\times$Crops$\times$Clips & \bfseries FLOPs & \bfseries \#Param. & \bfseries Top-1 (\%)  & \bfseries  Top-5 (\%) \\
      \midrule
      NL I3D GCN~\cite{wang2018videos} & \multirow{6}*{ImageNet} & 3D R50 & 32$\times$3$\times$2 & 303G$\times$3$\times$2 &62.2M & 46.1 & 76.8\\
      
      TRN$_{\textup{RGB+Flow}}$~\cite{zhou2018temporal} & & BNInception & (8$+$48)$\times$1$\times$1 & N/A & 36.6M& 42.0 & -\\

      TSM$_{\mathrm{En}}$~\cite{lin2019tsm} &  & R50  & (16$+$8)$\times$1$\times$1 & 98G & 48.6M&  49.7 & 78.5\\

      TEA~\cite{li2020tea}  & & R50 & 16$\times$3$\times$10 & 70G$\times$3$\times$10 & 24.4M& 52.3 & 81.9\\
    bLVNet-TAM~\cite{fan2019more}&  & bLR50 & 8$\times$1$\times$2 & 12G$\times$1$\times$2 & 25M & 46.4 & 76.6\\
    PAN$_{\mathrm{Full}}$~\cite{zhang2019pan} & & TSM R50 & 40$\times$1$\times$2 &67.7G$\times$1$\times$2 & - & 50.5 & 79.2\\
    ir-CSN~\cite{tran2019video} & None & 3D R152 & 32$\times$1$\times$10 &96.7G$\times$1$\times$10 & - & 49.3 & - \\
    \midrule
    TSM R50~\cite{lin2019tsm} & \multirow{5}*{ImageNet}  & R50 & 16$\times$1$\times$1 & 65G$\times$1$\times$1& 24.3M & 47.2 & 77.1\\
    \textbf{BQN}&  & TSM R50 & 24$\times$1$\times$1 & 60G$\times$1$\times$1& 47.4M & 51.7 & 80.5 \\
    \textbf{BQN}& &  TSM R50 & 24$\times$3$\times$2 & 60G$\times$3$\times$2& 47.4M &  53.3 & 82.0 \\
    \textbf{BQN}& &  TSM R50 & 48$\times$3$\times$2 & 121G$\times$3$\times$2& 47.4M & 54.3 & 82.0 \\
    \textbf{BQN}& & TSM R101 & 48$\times$3$\times$2 & 231G$\times$3$\times$2 &  85.4M& 54.9 & 81.7 \\
    \midrule
    X3D-M$\dagger$~\cite{feichtenhofer2020x3d} & None  & - & 16$\times$3$\times$2 &6.4G$\times$3$\times$2 & 3.3M &  46.7 & 75.5\\
    \textbf{BQN}& None  & X3D-M & 48$\times$3$\times$2 & 9.7G$\times$3$\times$2 & 6.6M & 50.6 & 79.2 \\
    \textbf{BQN}& K400 & X3D-M & 48$\times$3$\times$2 & 9.7G$\times$3$\times$2 & 6.6M & 53.7 & 81.8 \\
    \midrule
    \multirow{2}*{\textbf{BQN$_{\mathrm{En}}$}} & ImageNet & TSM R101 & \multirow{2}*{(48+48)$\times$3$\times$2} & \multirow{2}*{241G$\times$3$\times$2} & \multirow{2}*{92M} & \multirow{2}*{\textbf{57.1}} & \multirow{2}*{\textbf{84.2}} \\
    & + K400 & +X3D-M &  & & & & \\
    \bottomrule
      \end{tabular}
      \end{center}
       \vspace{-0.8cm}
\end{table*}

\noindent \textbf{BQN \vs Quiet+Busy.} 
In order to evaluate the effectiveness of the proposed BQN architecture, we compare BQN with the simple fusion (Quiet+Busy), which mimics the two-stream model~\cite{simonyan2014two} by averaging the predictions of two pathways trained separately.
Table~\ref{tab:abl_fcnet:fusion} shows that the simple fusion of two individual pathways (Quiet+Busy) generates higher top-1 accuracy (50.3\%) than the individual pathways, which indicates that the features learned by the Quiet pathway and by the Busy pathway are complementary. Surprisingly, BQN has 51.6\% top-1 accuracy, which is 1.3\% better than the Quiet+Busy fusion. The high-performance gain strongly demonstrates the advantages of the proposed BQN architecture.

\noindent \textbf{Fusion strategies} applied at the end of the Busy and Quiet pathways also influence the performance of BQN. Table~\ref{tab:abl_fcnet:fusion_strategy} shows the results of different fusions. We observe that the average fusion gives the best result among the listed methods, while the concatenation fusion is second only to the averaging. Besides, placing the average fusion layer after the fully-connected layer is better than placing it before.

\noindent \textbf{Effectiveness of the BPLC.} 
We can set a maximum of up to 16 BPLCs in the BQN architecture when using TSM R50~\cite{lin2019tsm} as the backbone\footnote{ResNet50 contains four stages, named res2, res3, res4, res5, respectively. These stages are composed of 3, 4, 6, 3 residual blocks, respectively.}. For the BPLCs in stage res2, res3 and res4, we set the spatial kernel size of MBPM as $7\times7$, and the scale $\sigma = 0.9$. As for the stage res5, whose feature size is relatively small, the kernel size is therefore set to $3\times3$.
Table~\ref{tab:abl_fcnet: lc_stage}, illustrates that adding BPLCs to all processing stages is helpful for improving performance.
From Table~\ref{tab:abl_fcnet:num_lc}, we can observe that the model performance improves gradually as the number of BPLCs increases. 
The substantial performance gains demonstrate the importance of using BPLCs for BQN.

\noindent \textbf{Lateral Connection (LC) Designs.}
In order to illustrate the rationality of the proposed BPLC design, we compare it with other LC designs.
The diagrams of different LC designs are illustrated in Figure~\ref{fig:lc_variant}, where LC-I and LC-II are unidirectional, and LC-III is bidirectional. The results from
Table~\ref{tab:abl_fcnet: lc_variants} show that the bidirectional design LC-III has higher accuracy than the unidirectional designs LC-I and LC-II. Among the listed designs, the proposed BPLC, which reverses the information fusion direction back and forth, provides the highest accuracy. We also compare the BPLC with LC-V that does not contain an MBPM. As a result, LC-V shows lower accuracy than the BPLC, which demonstrates the importance of MBPM for the BPLC.

\noindent \textbf{Spatial-temporal input size.} 
In BQN, the Busy pathway takes as input the MBPM output, which has the same spatial size as the raw video clip, while the temporal size is one-third of the raw video clip length.
Meanwhile, the Quiet pathway takes as input the complementary of the MBPM output, given by equation~\eqref{eq:quiet_input}.
Table~\ref{tab:abl_fcnet: size} shows that with the same temporal size of 8 for the inputs, the spatial size combination of $160^2$ and $256^2$ for the Quiet and Busy, respectively, has slightly better top-1 accuracy (+0.1\%) than the combination of $224^2$ and $224^2$ but saves 5.2 GFLOPs in computational cost. We also attempt to reduce the temporal input size of the Quiet pathway. However, this would result in a  performance drop. One possible explanation is that due to the temporal average pooling in the Quiet pathway, the input's temporal size is already reduced to one-third of the raw video clip. An even smaller temporal size could fail to preserve the correct temporal order of the video, and therefore harms the temporal relation modeling.
\begin{table}[t]
\small
     \caption{\label{tab:sota_kinetics} Comparison results on Kinetics400. We report the inference cost of multiple ``views'' (spatial crops × temporal clips). $\dagger$ denotes our reimplementation.}
     \vspace{-0.3cm}
      \begin{center}
      \setlength{\tabcolsep}{1.3pt}
      \begin{tabular}{l c c c c c c}
      \toprule
      \bfseries Method  &\bfseries Backbone &  \tabincell{c}{\bfseries Frames \\ \bfseries $\times$ views} &  \bfseries FLOPs & \tabincell{c}{\bfseries Top-1 \\ \bfseries (\%)} & \tabincell{c}{\bfseries Top-5 \\ \bfseries (\%)} \\
      \midrule
      bLVNet-TAM~\cite{fan2019more} & bLR50 & 16$\times$9 & 561G & 72.0 & 90.6\\
      TSM~\cite{lin2019tsm}& R50 & 16$\times$30 & 2580G & 74.7 & - \\

      STM~\cite{jiang2019stm}& R50 & 16$\times$30 & 2010G & 73.7 & 91.6\\
      X3D-M$\dagger$~\cite{feichtenhofer2020x3d} & - & 16$\times$30 & 186G & 75.1 & 92.2\\

      \midrule
      SlowFast$_{4\times16}$~\cite{feichtenhofer2019slowfast} & 3D R50 & 32$\times$30 & 1083G & 75.6 & 92.1\\

      ip-CSN~\cite{tran2019video} & 3D R101 & 32$\times$30& 2490G & 76.7 & 92.3\\
      SmallBigNet~\cite{li2020smallbignet} & R101 & 32$\times$12 & 6552G & 77.4 & 93.3\\

      \midrule
      PAN$_{\mathrm{Full}}$ & TSM R50 & 40$\times$2 & 176G & 74.4 & 91.6\\
      I3D$_{\mathrm{RGB}}$~\cite{carreira2017quo}  & Inc. V1 & 64$\times$N/A & N/A & 71.1 & 89.3\\
      Oct-I3D~\cite{chen2019drop}  & - & N/A$\times$N/A & N/A & 74.6 & - \\
      NL I3D~\cite{wang2018non} & 3D R101 & 128$\times$30 & 10770G & 77.7 & 93.3\\
      \midrule
      \textbf{BQN} & TSM R50 & 48$\times$10 & 1210G & 76.8 & 92.4 \\
      \textbf{BQN} & TSM R50 & 72$\times$10 & 1820G & 77.3 & 93.2 \\
      \textbf{BQN} & X3D-M & 48$\times$30 & 291G & 77.1 & 92.5 \\
    \bottomrule
      \end{tabular}
      \vspace{-0.4cm}
      \end{center}
\end{table}

\subsection{Comparisons with the State-of-the-Art}
We compare BQN with current state-of-the-art methods on the four datasets.
 In BQN, the Quiet and Busy pathways' spatial input size is set to $160^2$ and $256^2$, respectively. 

\noindent \textbf{Results on Something-Something V1.} Table~\ref{tab:sota_something} summarizes the comprehensive comparison, including the inference protocols, corresponding computational costs (FLOPs) and the prediction accuracy.
Our method surpasses all other methods by good margins. For example, the multi-clip accuracy of BQN$_{24f}$ with TSM R50 is 7.2\% higher than NL I3D GCN$_{32f}$~\cite{wang2018videos} while requiring $5\times$ fewer FLOPs. Among the models based on ResNet50, BQN$_{48f}$ has the highest top-1 accuracy (54.3\%), which surpasses the second-best, TEA$_{16f}$~\cite{li2020tea}, by a margin of +2\%. Furthermore, our signal-clip BQN$_{24f}$ has higher accuracy (51.7\%) than most other multi-clip models, requiring only 60 GFLOPs.
By adopting a deeper backbone (TSM R101), BQN$_{48f}$ has 54.9\% top-1 accuracy, higher than any single model. 
When using X3D-M as the backbone, BQN achieves the ultimate efficiency, possessing very low redundancy in both feature channel and spatio-temporal dimensions. BQN with X3D-M processes 4$\times$ more video frames than vanilla X3D-M, with only 50\% additional FLOPs. Compared with TSM R50$_{16f}$, BQN with X3D-M trained from scratch produces 3.4\% higher top-1 accuracy with the computation complexity of 14\% of TSM R50$_{16f}$.
The ensemble version \textbf{BQN$_{\mathrm{En}}$} achieves the state-of-the-art top-1/5 accuracy (57.1\%/84.2\%).
 \begin{table}[tp]
      \small
     \caption{\label{tab: ucf_hmdb} Results on HMDB51 and UCF101. We report the mean class accuracy over the three official splits.}
     \vspace{-0.3cm}
      \begin{center}
      \setlength{\tabcolsep}{4pt}
      \begin{tabular}{l c c c}
      \toprule
      \bfseries Method  & {\bfseries Backbone} & \bfseries HMDB51 &  \bfseries UCF101  \\ \midrule 
      StNet~\cite{he2019stnet} & R50  & - & 93.5 \\
      TSM~\cite{lin2019tsm} & R50  & 73.5 & 95.9 \\
      STM~\cite{jiang2019stm} & R50 & 72.2 & 96.2 \\ 
      TEA~\cite{li2020tea} & R50  & 73.3 & 96.9 \\
      DI Four-Stream~\cite{bilen2017action} & ResNeXt101  & 72.5 & 95.5 \\
      TVNet~\cite{fan2018end} & BNInception  & 71.0 & 94.5 \\
      TSN$_{\textup{RGB+Flow}}$~\cite{wang2016temporal} & BNInception & 68.5 & 94.0 \\
      I3D$_{\textup{RGB+Flow}}$~\cite{carreira2017quo} & 3D Inception &
      {\bfseries 80.7} & {\bfseries 98.0} \\
      PAN$_{\textup{Full}}$~\cite{zhang2019pan} & TSM R50  & 77.0 & 96.5 \\
      \midrule
        \textbf{BQN} & \textbf{TSM R50} & \textbf{77.6} & \textbf{97.6}\\
     \bottomrule
      \end{tabular}
      \vspace{-0.6cm}
      \end{center}
\end{table}

\noindent \textbf{Results on Kinetics400, UCF101 and HMDB51.} 
Table~\ref{tab:sota_kinetics} shows the comparison results on Kinetics400. For fair comparison, we only list the models with the spatial input size of $256^2$.
BQN$_{72f}$ with TSM R50 achieves 77.3\%/93.2\% top-1/5 accuracy, which is better than the 3D CNN-based architecture, I3D~\cite{carreira2017quo}, by a big margin of +6.2\%/3.9\%. 
When BQN uses TSM R50 or X3D-M as the backbone, it consistently shows higher accuracy than SlowFast$_{4\times16}$. Particularly, BQN with X3D-M has 1.5\% higher top-1 accuracy than SlowFast$_{4\times16}$, while requiring $3.7\times$ fewer FLOPs. Meanwhile, BQN$_{72f}$ with TSM R50 is 2.7\% better than Oct-I3D~\cite{chen2019drop} for top-1 accuracy. 
The results on two smaller datasets, UCF101 and HMDB51, are shown in Table~\ref{tab: ucf_hmdb}, where we report the mean class accuracy over the three official splits.  
We pretrain our model on Kinetics400 to avoid overfitting. The accuracy of our method is obtained by the inference protocol (3 crops$\times$2 clips). BQN with TSM R50 outperforms most other methods except for I3D$_{\textup{RGB+Flow}}$, which uses additional optical flow input modality. 

\vspace*{-0.2cm}
\section{Conclusion}
\vspace*{-0.1cm}

This paper develops a novel video representation learning mechanism called Motion Band-Pass Module (MBPM). The MBPM can distill important motion cues corresponding to a set of band-pass spatio-temporal frequencies. We design a spatio-temporal architecture called Bus-Quiet Net (BQN), enabled by MBPM, to separately process busy and quiet video data information. The busy-quiet disentangling enables efficient video processing by allocating additional resources to the Busy stream and less to the Quiet. Our methods can also be used for video processing and analysis in various applications.

\vspace*{-0.2cm}
\section*{Acknowledgments}
\vspace*{-0.1cm}
This work made use of the facilities of the N8 Centre of Excellence in Computationally Intensive Research (N8 CIR) provided and funded by the N8 research partnership and EPSRC (Grant No. EP/T022167/1).
 
{\small
\bibliographystyle{ieee_fullname}
\bibliography{egpaper.bbl}

\begin{thebibliography}{10}\itemsep=-1pt

\bibitem{baker2011database}
S. Baker, D. Scharstein, J. Lewis, S. Roth, M.~J. Black, and R. Szeliski.
\newblock A database and evaluation methodology for optical flow.
\newblock In {\em Proc. IEEE Int. Conference Computer Vision (ICCV)}, pages
  1--8, 2007.

\bibitem{bilen2017action}
H. Bilen, B. Fernando, E. Gavves, and A. Vedaldi.
\newblock Action recognition with dynamic image networks.
\newblock {\em IEEE Trans. on Pattern Analysis and Machine Intelligence},
  40(12):2799--2813, 2018.

\bibitem{cao2019gcnet}
Y. Cao, J. Xu, S. Lin, F. Wei, and H. Hu.
\newblock Gcnet: Non-local networks meet squeeze-excitation networks and
  beyond.
\newblock In {\em Proc. IEEE Int. Conference Computer Vision Workshops
  (ICCV-w)}, 2019.

\bibitem{carreira2017quo}
J. Carreira and A. Zisserman.
\newblock Quo vadis, action recognition? a new model and the kinetics dataset.
\newblock In {\em Proc. IEEE Conference Computer Vision Pattern Recog. (CVPR)},
  pages 4724--4733, 2017.

\bibitem{chen2018big}
C. Chen, Q. Fan, N. Mallinar, T. Sercu, and R. Feris.
\newblock Big-little net: An efficient multi-scale feature representation for
  visual and speech recognition.
\newblock {\em arXiv preprint arXiv:1807.03848}, 2018.

\bibitem{chen2019drop}
Y. Chen, H. Fan, B. Xu, Z. Yan, Y. Kalantidis, M. Rohrbach, S. Yan, and J.
  Feng.
\newblock Drop an octave: Reducing spatial redundancy in convolutional neural
  networks with octave convolution.
\newblock In {\em Proc. IEEE Int. Conference Computer Vision (ICCV)}, pages
  3435--3444, 2019.

\bibitem{imagenet_cvpr09}
J. Deng, W. Dong, R. Socher, L.-J. Li, K. Li, and L. Fei-Fei.
\newblock Imagenet: A large-scale hierarchical image database.
\newblock In {\em Proc. IEEE Conference Computer Vision Pattern Recog. (CVPR)},
  pages 248--255, 2009.

\bibitem{donahue2015long}
J. Donahue, L. Anne~Hendricks, S. Guadarrama, M. Rohrbach, S. Venugopalan, K.
  Saenko, and T. Darrell.
\newblock Long-term recurrent convolutional networks for visual recognition and
  description.
\newblock In {\em Proc. IEEE Conference Computer Vision Pattern Recog. (CVPR)},
  pages 2625--2634, 2015.

\bibitem{dosovitskiy2015flownet}
A. Dosovitskiy, P. Fischer, E. Ilg, P. Hausser, C. Hazirbas, V. Golkov, P. Van
  Der~Smagt, D. Cremers, and T. Brox.
\newblock Flownet: Learning optical flow with convolutional networks.
\newblock In {\em Proc. IEEE Int. Conf. Computer Vision (ICCV)}, pages
  2758--2766, 2015.

\bibitem{fan2018end}
L. Fan, W. Huang, C. Gan, S. Ermon, B. Gong, and J. Huang.
\newblock End-to-end learning of motion representation for video understanding.
\newblock In {\em Proc. IEEE Conference Computer Vision Pattern Recog. (CVPR)},
  pages 6016--6025, 2018.

\bibitem{fan2019more}
Q. Fan, C. Chen, H. Kuehne, M. Pistoia, and D. Cox.
\newblock More is less: Learning efficient video representations by big-little
  network and depthwise temporal aggregation.
\newblock In {\em Advances Neural Information Process. Systems (NIPS)}, pages
  2264--2273, 2019.

\bibitem{feichtenhofer2020x3d}
C Feichtenhofer.
\newblock X3d: Expanding architectures for efficient video recognition.
\newblock In {\em Proc. IEEE Conference Computer Vision Pattern Recog. (CVPR)},
  pages 203--213, 2020.

\bibitem{feichtenhofer2019slowfast}
C. Feichtenhofer, H. Fan, J. Malik, and K. He.
\newblock Slowfast networks for video recognition.
\newblock In {\em Proc. IEEE Conf. Computer Vision Pattern Recog. (CVPR)},
  pages 6202--6211, 2019.

\bibitem{feichtenhofer2016spatiotemporal}
C. Feichtenhofer, A. Pinz, and R. Wildes.
\newblock Spatiotemporal residual networks for video action recognition.
\newblock In {\em Advances in Neural Information Processing Systems (NIPS)},
  pages 3468--3476, 2016.

\bibitem{feichtenhofer2016convolutional}
C. Feichtenhofer, A. Pinz, and A. Zisserman.
\newblock Convolutional two-stream network fusion for video action recognition.
\newblock In {\em Proc. IEEE Conference Computer Vision Pattern Recog. (CVPR)},
  pages 1933--1941, 2016.

\bibitem{goyal2017something}
R. Goyal, S.~E. Kahou, V. Michalski, J. Materzynska, S. Westphal, H. Kim, V.
  Haenel, I. Fruend, P. Yianilos, M. Mueller-Freitag, et~al.
\newblock The" something something" video database for learning and evaluating
  visual common sense.
\newblock In {\em Proc. IEEE Int. Conf. on Computer Vision (ICCV)}, volume~1,
  pages 5842--5850, 2017.

\bibitem{hara2018can}
K. Hara, H. Kataoka, and Y. Satoh.
\newblock Can spatiotemporal {3D CNNs} retrace the history of {2D CNNs} and
  {ImageNet}?
\newblock In {\em Proc. IEEE Conference Computer Vision Pattern Recog. (CVPR)},
  pages 6546--6555, 2018.

\bibitem{he2019stnet}
D. He, Z. Zhou, C. Gan, F. Li, X. Liu, Y. Li, L. Wang, and S. Wen.
\newblock Stnet: Local and global spatial-temporal modeling for action
  recognition.
\newblock In {\em Proc. AAAI Conference on Artif. Intel.}, volume~33, pages
  8401--8408, 2019.

\bibitem{he2016deep}
K. He, X. Zhang, S. Ren, and J. Sun.
\newblock Deep residual learning for image recognition.
\newblock In {\em Proc. IEEE Conference Computer Vision Pattern Recog. (CVPR)},
  pages 770--778, 2016.

\bibitem{huang2020learning}
G. Huang and A.~G. Bors.
\newblock Learning spatio-temporal representations with temporal squeeze
  pooling.
\newblock In {\em Proc. IEEE Int. Conference on Acoustics, Speech and Signal
  Processing (ICASSP)}, pages 2103--2107, 2020.

\bibitem{huang2021icpr}
G. Huang and A.~G. Bors.
\newblock Region-based non-local operation for video classification.
\newblock In {\em Proc. Int. Conference on Pattern Recognition (ICPR)}, pages
  10010--10017, 2021.

\bibitem{huang2017densely}
G. Huang, Z. Liu, L. Van Der~Maaten, and K.~Q Weinberger.
\newblock Densely connected convolutional networks.
\newblock In {\em Proc. IEEE Conference Computer Vision Pattern Recog. (CVPR)},
  pages 4700--4708, 2017.

\bibitem{ilg2017flownet}
E. Ilg, N. Mayer, T. Saikia, M. Keuper, A. Dosovitskiy, and T. Brox.
\newblock Flownet 2.0: Evolution of optical flow estimation with deep networks.
\newblock In {\em Proc. IEEE Conference Computer Vision Pattern Recog. (CVPR)},
  volume~2, pages 2462--2470, 2017.

\bibitem{szegedy2015bn}
S. Ioffe and C. Szegedy.
\newblock Batch normalization: Accelerating deep network training by reducing
  internal covariate shift.
\newblock In {\em Proc. Int. Conference Mach. Learn. (ICML), vol. PMLR 37},
  page 448–456, 2015.

\bibitem{jiang2019stm}
B. Jiang, M. Wang, W. Gan, W. Wu, and J. Yan.
\newblock Stm: Spatiotemporal and motion encoding for action recognition.
\newblock In {\em Proc. IEEE Int. Conference Computer Vision (ICCV)}, pages
  2000--2009, 2019.

\bibitem{krizhevsky2012imagenet}
A. Krizhevsky, I. Sutskever, and G Hinton.
\newblock Imagenet classification with deep convolutional neural networks.
\newblock In {\em Advances Neural Information Process. Systems (NIPS)}, pages
  1097--1105, 2012.

\bibitem{kuehne2011hmdb}
H. Kuehne, H. Jhuang, E. Garrote, T. Poggio, and T. Serre.
\newblock {HMDB}: a large video database for human motion recognition.
\newblock In {\em Proc. IEEE Int. Conference Computer Vision (ICCV)}, pages
  2556--2563, 2011.

\bibitem{li2020smallbignet}
X. Li, Y. Wang, Z. Zhou, and Y. Qiao.
\newblock Smallbignet: Integrating core and contextual views for video
  classification.
\newblock In {\em Proc. IEEE Conference Computer Vision Pattern Recog. (CVPR)},
  pages 1092--1101, 2020.

\bibitem{li2020tea}
Y. Li, B. Ji, X. Shi, J. Zhang, B. Kang, and L. Wang.
\newblock Tea: Temporal excitation and aggregation for action recognition.
\newblock In {\em Proc. IEEE Conference Computer Vision Pattern Recog. (CVPR)},
  pages 909--918, 2020.

\bibitem{lin2019tsm}
J. Lin, C. Gan, and S. Han.
\newblock Tsm: Temporal shift module for efficient video understanding.
\newblock In {\em Proc. IEEE Int. Conference Computer Vision (ICCV)}, pages
  7083--7093, 2019.

\bibitem{lin2017feature}
T. Lin, P. Doll{\'a}r, R. Girshick, K. He, B. Hariharan, and S. Belongie.
\newblock Feature pyramid networks for object detection.
\newblock In {\em Proc. IEEE Conference Computer Vision Pattern Recog. (CVPR)},
  pages 2117--2125, 2017.

\bibitem{loshchilov2016sgdr}
I. Loshchilov and F. Hutter.
\newblock Sgdr: Stochastic gradient descent with warm restarts.
\newblock {\em Int. Conference Learn. Representations (ICLR), arXiv preprint
  arXiv:1608.03983}, 2017.

\bibitem{piergiovanni2019representation}
A. Piergiovanni and S. Ryoo.
\newblock Representation flow for action recognition.
\newblock In {\em Proc. IEEE Conference Computer Vision Pattern Recog. (CVPR)},
  pages 9945--9953, 2019.

\bibitem{qiu2017learning}
Z. Qiu, T. Yao, and T. Mei.
\newblock Learning spatio-temporal representation with pseudo-3d residual
  networks.
\newblock In {\em Proc. IEEE Int. Conference Computer Vision (ICCV)}, pages
  5533--5541, 2017.

\bibitem{sandler2018mobilenetv2}
M. Sandler, A. Howard, M. Zhu, A. Zhmoginov, and L. Chen.
\newblock Mobilenetv2: Inverted residuals and linear bottlenecks.
\newblock In {\em Proc. IEEE Conference Computer Vision Pattern Recog. (CVPR)},
  pages 4510--4520, 2018.

\bibitem{simonyan2014two}
K. Simonyan and A. Zisserman.
\newblock Two-stream convolutional networks for action recognition in videos.
\newblock In {\em Advances Neural Information Process. Systems (NIPS)}, pages
  568--576, 2014.

\bibitem{simonyan2014very}
K. Simonyan and A. Zisserman.
\newblock Very deep convolutional networks for large-scale image recognition.
\newblock In {\em Int. Conference Learn. Representations (ICLR), arXiv preprint
  arXiv:1409.1556}, 2015.

\bibitem{soomro2012ucf101}
K. Soomro, Amir~R. Zamir, and M. Shah.
\newblock {UCF101}: A dataset of 101 human actions classes from videos in the
  wild.
\newblock {\em arXiv preprint arXiv:1212.0402}, 2012.

\bibitem{sun2018optical}
S. Sun, Z. Kuang, L. Sheng, W. Ouyang, and W. Zhang.
\newblock Optical flow guided feature: A fast and robust motion representation
  for video action recognition.
\newblock In {\em Proc. IEEE Conference Computer Vision Pattern Recog. (CVPR)},
  pages 1390--1399, 2018.

\bibitem{szegedy2015going}
C. Szegedy, W. Liu, Y. Jia, P. Sermanet, S. Reed, D. Anguelov, D. Erhan, V.
  Vanhoucke, and A. Rabinovich.
\newblock Going deeper with convolutions.
\newblock In {\em Proc. IEEE Conference Computer Vision Pattern Recog. (CVPR)},
  pages 1--9, 2015.

\bibitem{tan2019efficientnet}
M. Tan and Q.~V Le.
\newblock Efficientnet: Rethinking model scaling for convolutional neural
  networks.
\newblock In {\em Proc. Int. Conference Mach. Learn. (ICML), vol. PMLR 97},
  pages 6105--6114, 2019.

\bibitem{taylor2010convolutional}
G.~W Taylor, R. Fergus, Y. LeCun, and C. Bregler.
\newblock Convolutional learning of spatio-temporal features.
\newblock In {\em Proc. European Conference Computer Vision (ECCV)}, pages
  140--153, 2010.

\bibitem{tran2015learning}
D. Tran, L. Bourdev, R. Fergus, L. Torresani, and M. Paluri.
\newblock Learning spatiotemporal features with {3D} convolutional networks.
\newblock In {\em Proc. IEEE Int. Conference Computer Vision (ICCV)}, pages
  4489--4497, 2015.

\bibitem{tran2019video}
D. Tran, H. Wang, L. Torresani, and M. Feiszli.
\newblock Video classification with channel-separated convolutional networks.
\newblock In {\em Proc. IEEE Int. Conference Computer Vision (ICCV)}, pages
  5552--5561, 2019.

\bibitem{tran2018closer}
D. Tran, H. Wang, L. Torresani, J. Ray, Y. LeCun, and M. Paluri.
\newblock A closer look at spatiotemporal convolutions for action recognition.
\newblock In {\em Proc. IEEE Conference Computer Vision Pattern Recog. (CVPR)},
  pages 6450--6459, 2018.

\bibitem{wang2018video}
J. Wang, A. Cherian, F. Porikli, and S. Gould.
\newblock Video representation learning using discriminative pooling.
\newblock In {\em Proc. IEEE Conference Computer Vision Pattern Recog. (CVPR)},
  pages 1149--1158, 2018.

\bibitem{wang2016temporal}
L. Wang, Y. Xiong, Z. Wang, Y. Qiao, D. Lin, X. Tang, and L. Van~Gool.
\newblock Temporal segment networks: Towards good practices for deep action
  recognition.
\newblock In {\em Proc. European Conference Computer Vision (ECCV), vol LNCS
  9912}, pages 20--36, 2016.

\bibitem{wang2018non}
X. Wang, R. Girshick, A. Gupta, and K. He.
\newblock Non-local neural networks.
\newblock In {\em Proc. IEEE Conference Computer Vision Pattern Recog. (CVPR)},
  pages 7794--7803, 2018.

\bibitem{wang2018videos}
X. Wang and A. Gupta.
\newblock Videos as space-time region graphs.
\newblock In {\em Proc. European Conference Computer Vision (ECCV)}, pages
  399--417, 2018.

\bibitem{xie2018rethinking}
S. Xie, C. Sun, J. Huang, Z. Tu, and K. Murphy.
\newblock Rethinking spatiotemporal feature learning: Speed-accuracy trade-offs
  in video classification.
\newblock In {\em Proc. European Conference Computer Vision (ECCV), vol. LNCS
  11219}, pages 305--321, 2018.

\bibitem{yue2015beyond}
J. Yue-Hei~Ng, M. Hausknecht, S. Vijayanarasimhan, O. Vinyals, R. Monga, and G.
  Toderici.
\newblock Beyond short snippets: Deep networks for video classification.
\newblock In {\em Proc. IEEE Conference Computer Vision Pattern Recog. (CVPR)},
  pages 4694--4702, 2015.

\bibitem{zach2007duality}
C. Zach, T. Pock, and H. Bischof.
\newblock A duality based approach for realtime tv-l 1 optical flow.
\newblock In {\em Proc. Joint Pattern Recog. Symp., vol. LNCS 4713}, pages
  214--223, 2007.

\bibitem{zhang2018real}
B. Zhang, L. Wang, Z. Wang, Y. Qiao, and H. Wang.
\newblock Real-time action recognition with deeply transferred motion vector
  cnns.
\newblock {\em IEEE Transactions on Image Processing}, 27(5):2326--2339, 2018.

\bibitem{zhang2019pan}
C. Zhang, Y. Zou, G. Chen, and L. Gan.
\newblock Pan: Persistent appearance network with an efficient motion cue for
  fast action recognition.
\newblock In {\em Proc. ACM Int. Conference Multimedia}, pages 500--509, 2019.

\bibitem{zhou2018temporal}
B. Zhou, A. Andonian, A. Oliva, and A. Torralba.
\newblock Temporal relational reasoning in videos.
\newblock In {\em Proc. European Conference Computer Vision (ECCV), vol LNCS
  11205}, pages 803--818, 2018.

\end{thebibliography}
}

\newpage

\appendix
\thispagestyle{empty}
\section{More training details}
\label{appen_a}
We train our models in 16 or 64 GPUs (NVIDIA Tesla V100), using Stochastic Gradient Descent (SGD) with momentum 0.9 and cosine learning rate schedule. In order to prevent overfitting, we add a dropout layer before the classification layer of each pathway in the BQN model. Following the experimental settings in~\cite{lin2019tsm,wang2016temporal}, the learning rate and weight decay parameters for the classification layers are 5 times of the convolutional layers. Meanwhile, we only apply L2 regularization to the weights in the convolutional and classification layers to avoid overfitting.

\noindent \textbf{Hyperparameters for models based on ResNet.}
For Kinetics400~\cite{carreira2017quo}, the initial learning rate, batch size, total epochs, weight decay and dropout ratio are set to 0.08, 512 (8 samples per GPU), 100, 2e-4 and 0.5, respectively. For Something-Something V1~\cite{goyal2017something}, these hyperparameters are set to 0.12, 256, 50, 8e-4 and 0.8, respectively. We use linear warm-up~\cite{loshchilov2016sgdr} for the first 7 epochs to overcome early optimization difficulty.
When fine-tuning the Kinetics models on UCF101~\cite{soomro2012ucf101} and HMDB51~\cite{kuehne2011hmdb}, we freeze all of the batch normalization~\cite{szegedy2015bn} layers except for the first one to avoid overfitting, following the recipe in~\cite{wang2016temporal}.
The initial learning rate, batch size, total epochs, weight decay and dropout ratio are set to 0.001, 64 (4 samples per GPU), 10, 1e-4 and 0.8, respectively.

\noindent \textbf{Hyperparameters for models based on X3D-M.}
For Kinetics400, the initial learning rate, batch size, total epochs, weight decay and dropout ratio are set to 0.4, 256 (16 samples per GPU), 256, 5e-5 and 0.5, respectively.
For Something-Something V1, the models trained from scratch use the followings hyperparameters: learning rate 0.2, batch size 256, total epochs 100, weight decay 5e-5 and dropout ratio 0.5.
When fine-tuning the Kinetics models, the initial learning rate, batch size, total epochs, weight decay and dropout ratio are set to 0.12, 256 (16 samples per GPU), 60, 4e-4 and 0.8, respectively.

\section{More Studies for the MBPM settings}
\begin{figure}[!t]
    \centering
    \includegraphics[width=.93\linewidth]{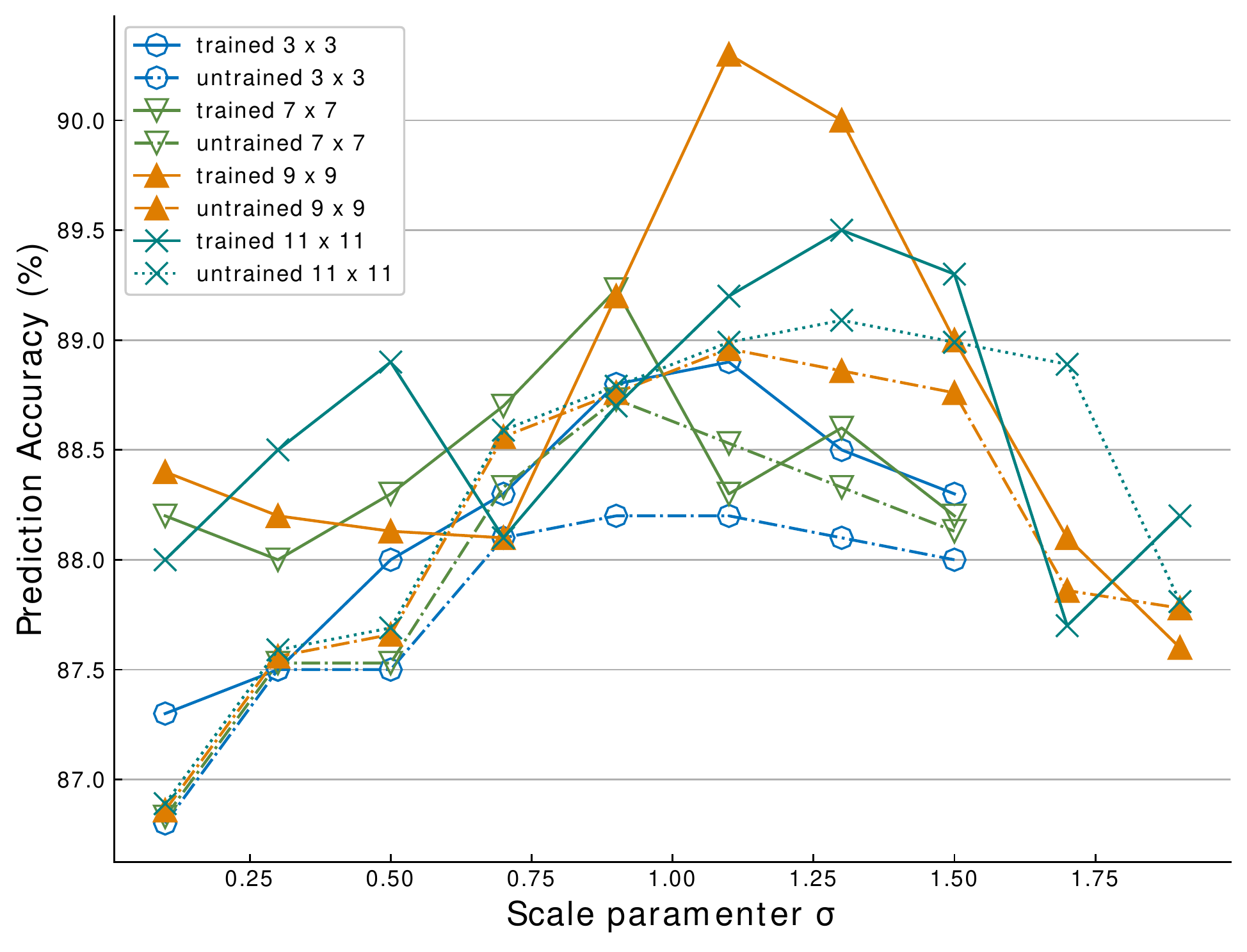}
    \vspace{-6pt}
    \caption{Results on UCF101 when varying the  scale $\sigma$ and kernel size $k \times k$ of the spatial channel-wise convolution in MBPM.}
    \label{fig:sima_eva_ucf}
\end{figure}
We search for the optimal settings of the scale $\sigma$ and kernel size $k \times k$ of MBPM on UCF101. The results are presented in Figure~\ref{fig:sima_eva_ucf}. We observe that the experimental results vary greatly under different settings. Nevertheless, the optimal scale is $\sigma=1.1$ when setting the kernel size as $9\times9$, which is the same as that on Something-Something dataset. Furthermore, we try a larger kernel ($11\times11$), but it shows a performance drop. We speculate that this is caused by insufficient training.

\section{Implementation details of Efficiency and Effectiveness of the MBPM}
These additional explanations are useful for Section 5.2 from the main paper.
We provide the implementation details for the comparative experiments of MBPM with other mainstream motion representation methods~\cite{bilen2017action,fan2018end,ilg2017flownet,wang2016temporal,zach2007duality,zhang2019pan}. We follow the experimental settings on PA~\cite{zhang2019pan} for fair comparison. The backbone network for all the methods is ResNet50~\cite{he2016deep}. We use the computer code provided by the original authors for these methods to generate the network inputs. For any kind of motion representation, we divide the representation of a video into 8 segments and randomly select one frame of the representation for each segment. Following the practices in TSN~\cite{wang2016temporal} and PA~\cite{zhang2019pan}, the activation outputs of 8 segments are averaged for the final prediction score. In our reimplementation, Dynamic Image~\cite{bilen2017action} generates one dynamic image for every 6 consecutive RGB frames, which consumes the same number of RGB frames as PA~\cite{zhang2019pan}. Our MBPM generates one representative frame for every 3 consecutive RGB frames. As for TVNet~\cite{fan2018end} and TV-L1 Flow~\cite{zach2007duality}, a one-frame input to the backbone network is formed by stacking 5 frames of the estimated flow along the channel dimension, which totally consumes 6 RGB frames. All the models are pretrained on ImageNet. For Something-Something V1 and Kinetics400, we use the hyperparameters in Appendix~\ref{appen_a} to train all the models. For UCF101, we set the initial learning rate, batch size, total epochs, weight decay and dropout ratio to 0.01, 64 (4 samples per GPU), 80, 1e-4 and 0.5, respectively.
\begin{figure*}[t]
    \centering
    \resizebox{1.\textwidth}{!}{\includegraphics{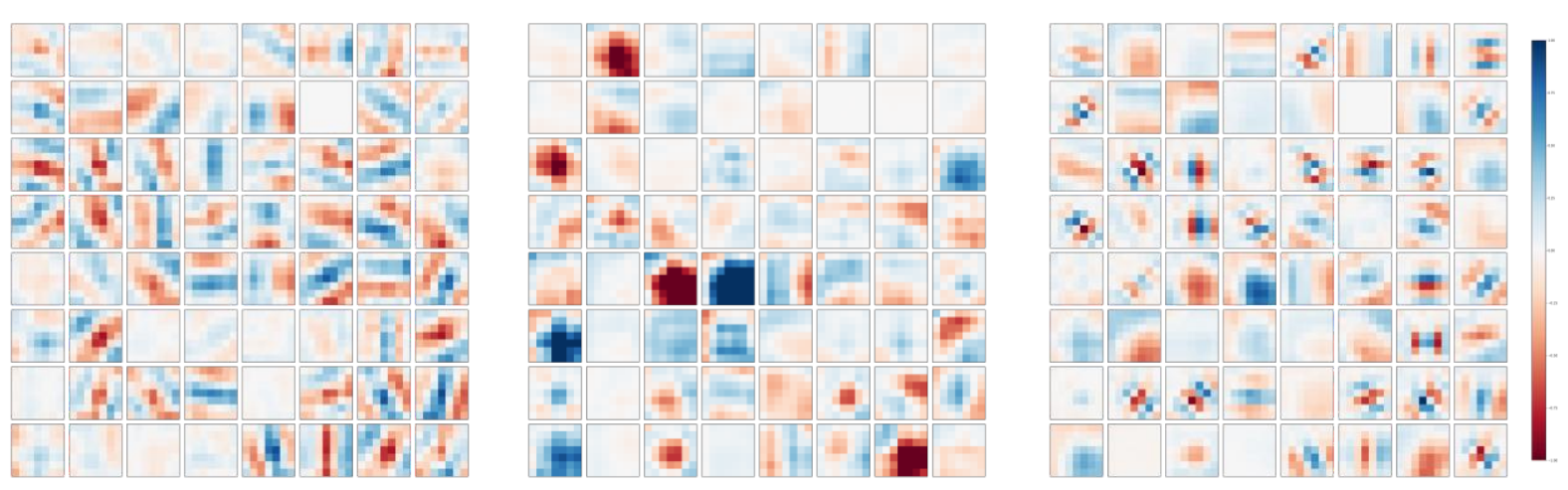}}
    (a) Busy pathway   \hspace{3.1cm} (b) Quiet pathway   \hspace{3.1cm} (c) TSM ResNet50
    \caption{Visualization of the first channels of the 64 conv1 filters of BQN after training on Kinetics400. All the 64 filters have a size of $7 \times 7$.
    From left to right, in (a), (b) and (c), we respectively present the trained conv1 filters in the Busy pathway, Quiet pathway and TSM ResNet50. We observe that the kernels of the 64 filters in the Busy pathway have a similar line-like shape, while those for the filters in the Quiet pathway are more like larger blobs. The conv1 in TSM ResNet50 (baseline) contains both types of filters from the Busy and Quiet pathways. 
    Best viewed in color and zoomed in.}
    \label{fig: visual_conv1}
\end{figure*}

\begin{figure}[b]
    \includegraphics[width=8.39cm]{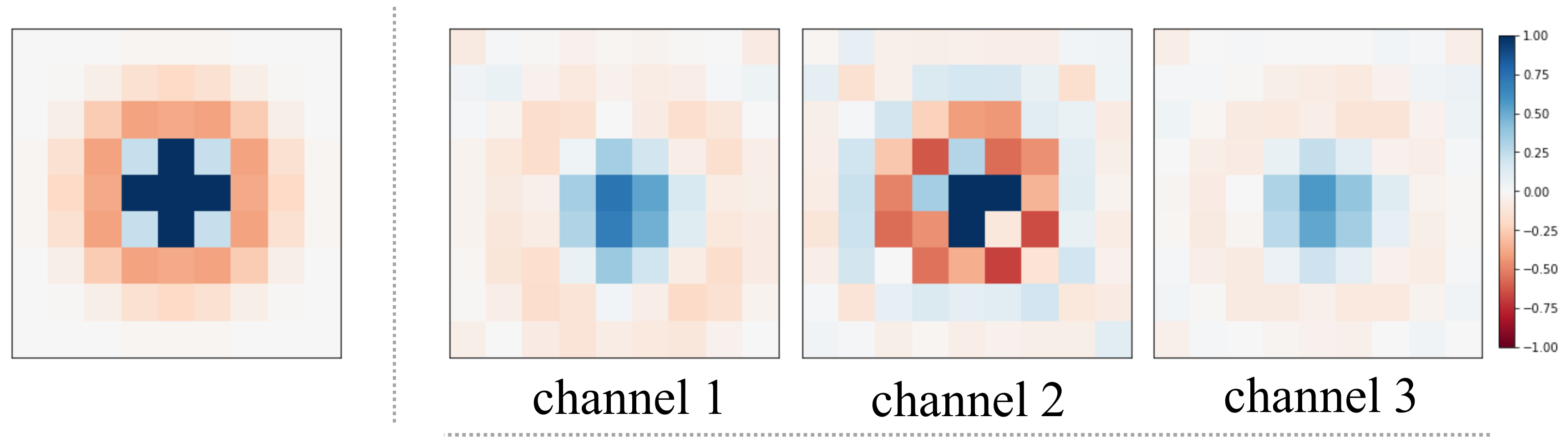}
    (a) untrained LoG \hspace{2cm} (b) trained LoG 
    \caption{Visualization of the spatial channel-wise convolution $LoG_\sigma^{1 \times k \times k}$ of MBPM in the Busy pathway before and after training on Kinetics400. The $9\times9$ channel-wise convolution is initialized with a Laplacian of Gaussian with the scale parameter $\sigma$~=~1.1. Best viewed in color and zoomed in.}
    \label{fig: mbpm_kernel}
\end{figure}

\begin{figure}[!t]
    \centering
    \includegraphics[width=8.5cm]{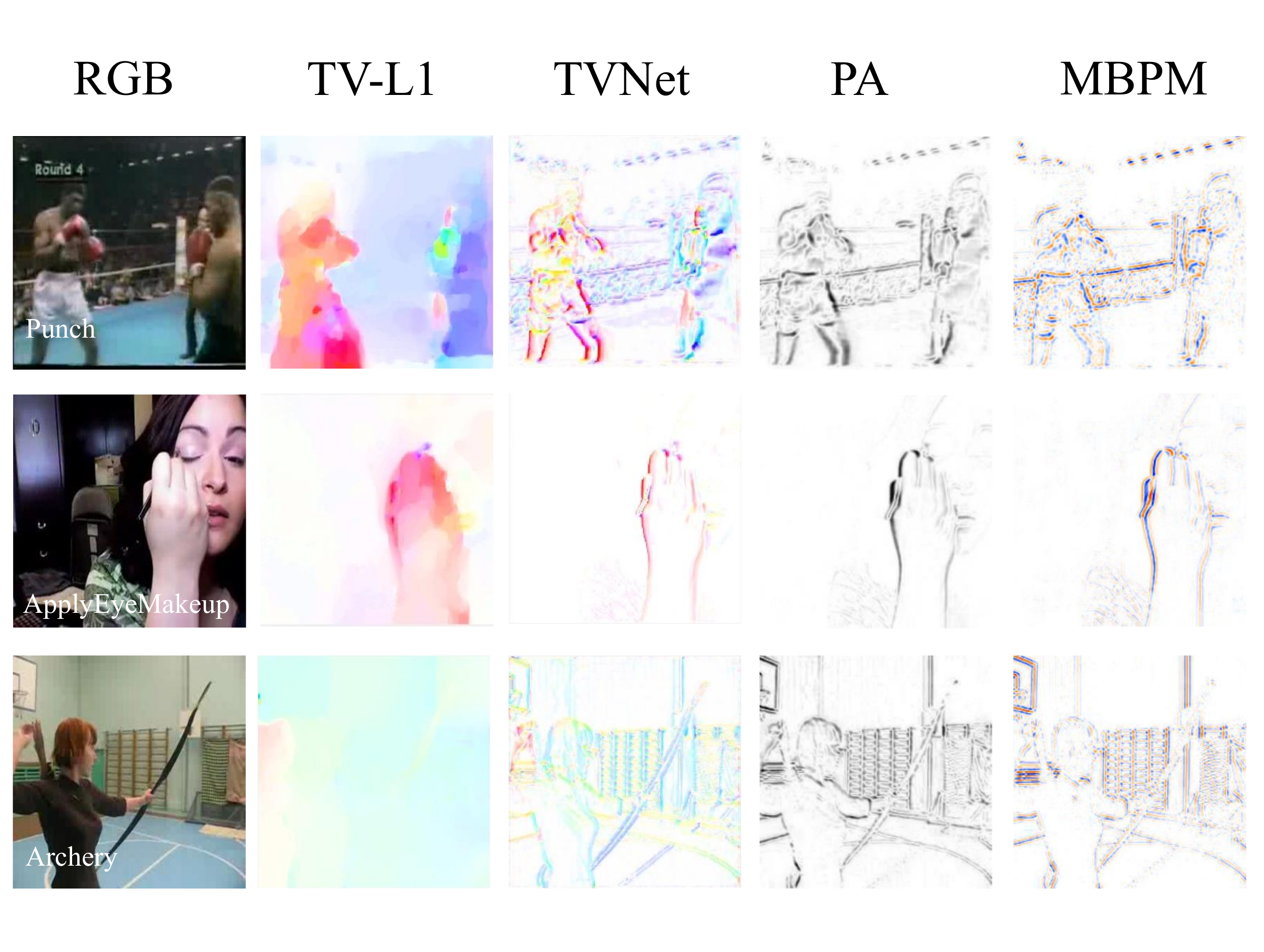}
    \caption{Comparison between visualizations of different motion representations on the UCF101. TV-L1 Flow~\cite{zach2007duality} evaluates the movement in every spatial position, while TVNet~\cite{fan2018end}, PA~\cite{zhang2019pan} and our MBPM capture the outline of the moving objects. Best viewed in color and zoomed in.}
    \label{fig:cmpar_other_reps}
    \vspace{-0.1cm}
\end{figure}

\section{Visualization examples}

\begin{figure*}[!t]
    \centering
    \resizebox{1.\textwidth}{!}{\includegraphics{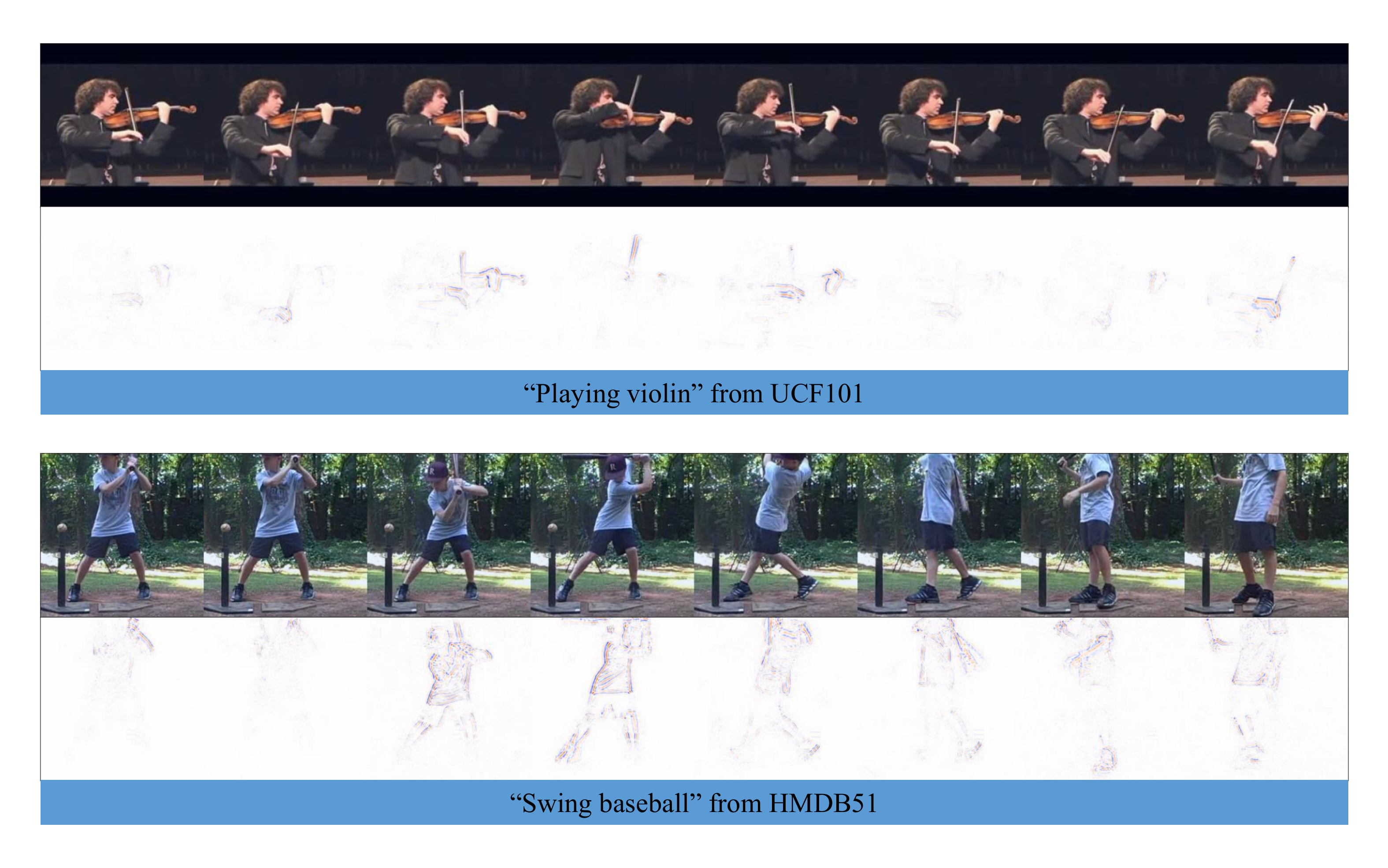}}
    \resizebox{1.\textwidth}{!}{\includegraphics{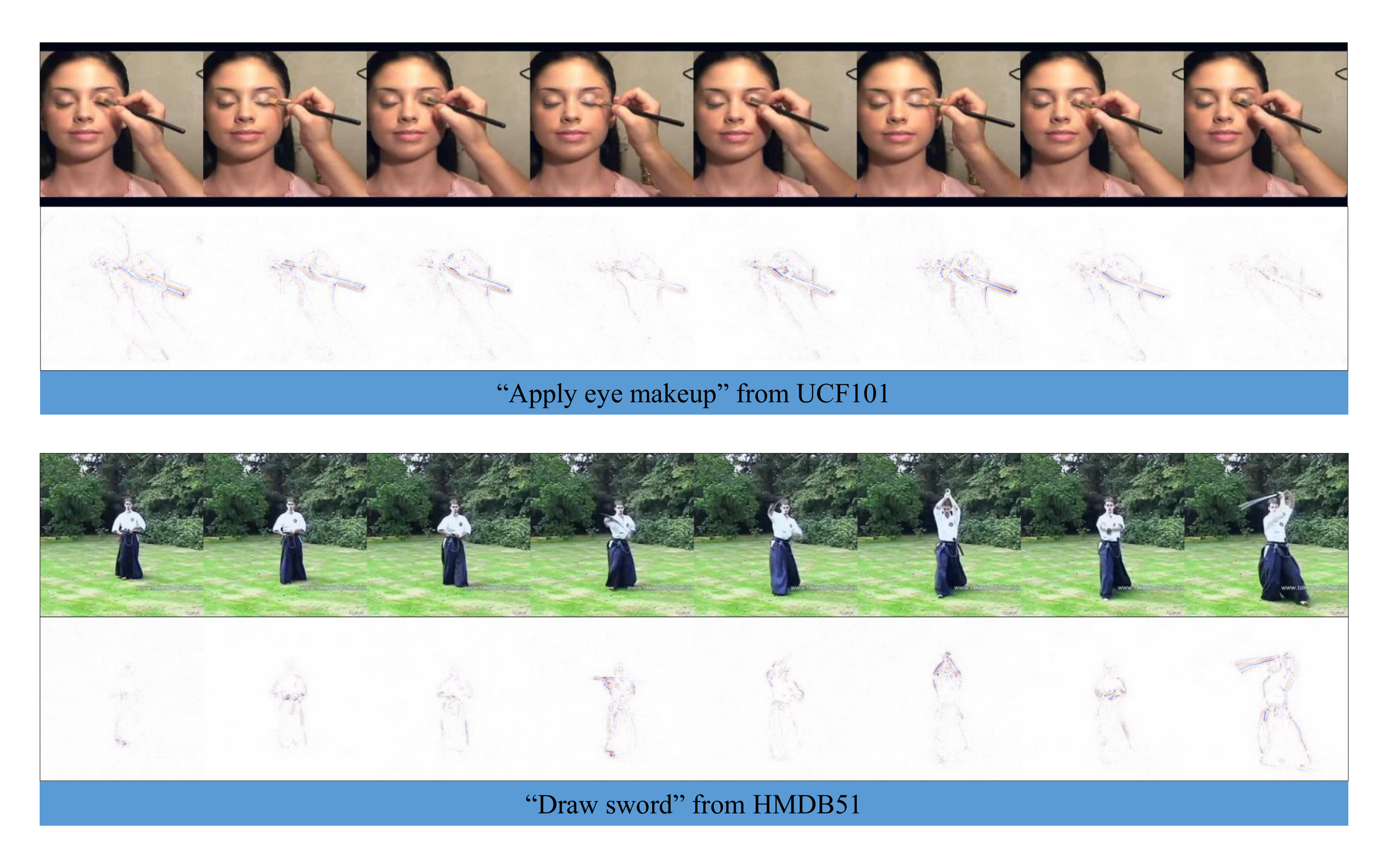}}
     \vspace*{-0.7cm}
    \caption{Examples of video and the corresponding motion representations extracted by MBPM.}
    \label{fig: more1}
\end{figure*}

\begin{figure*}[!t]
    \centering
    \resizebox{1.\textwidth}{!}{\includegraphics{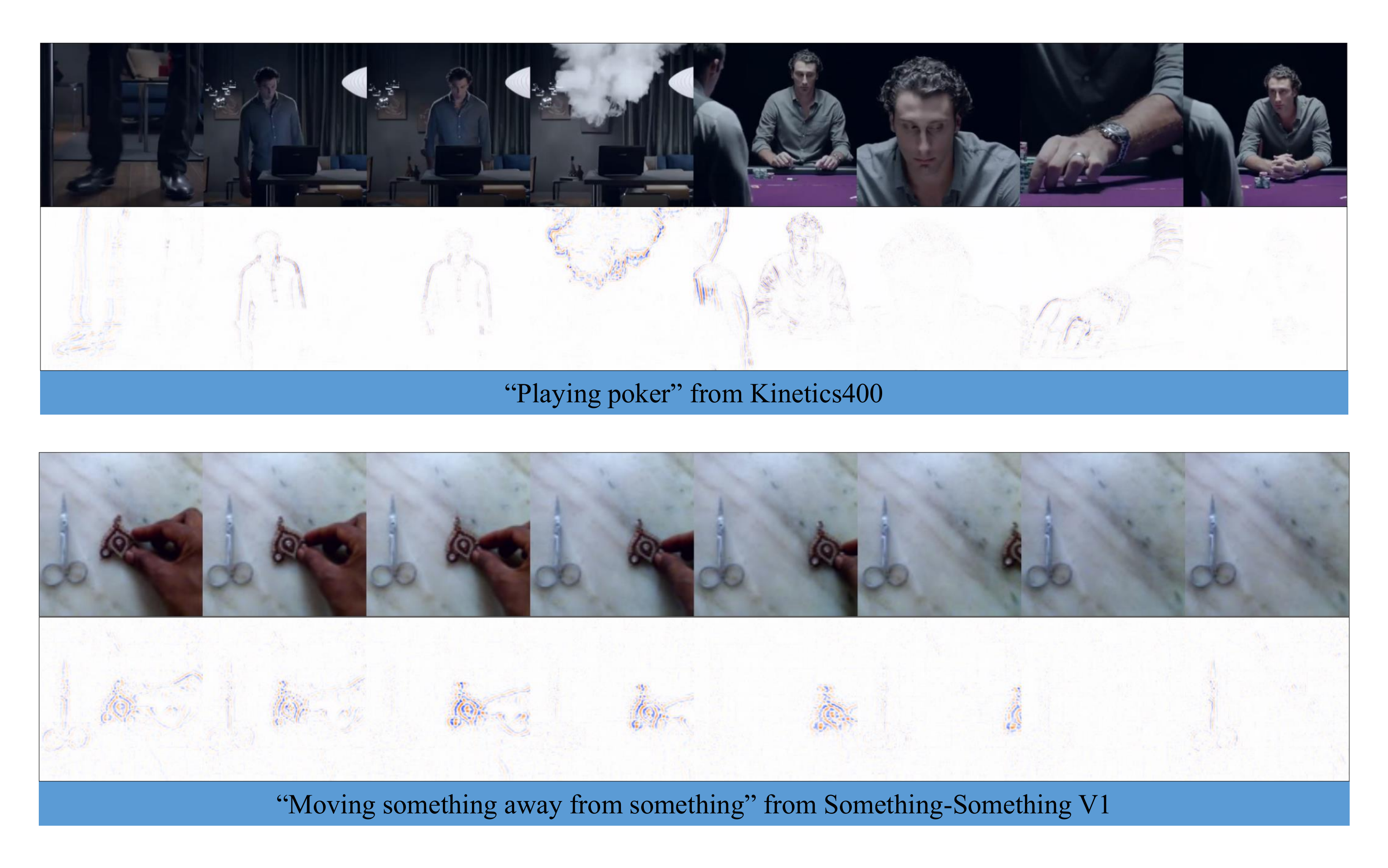}}
    \resizebox{1.\textwidth}{!}{\includegraphics{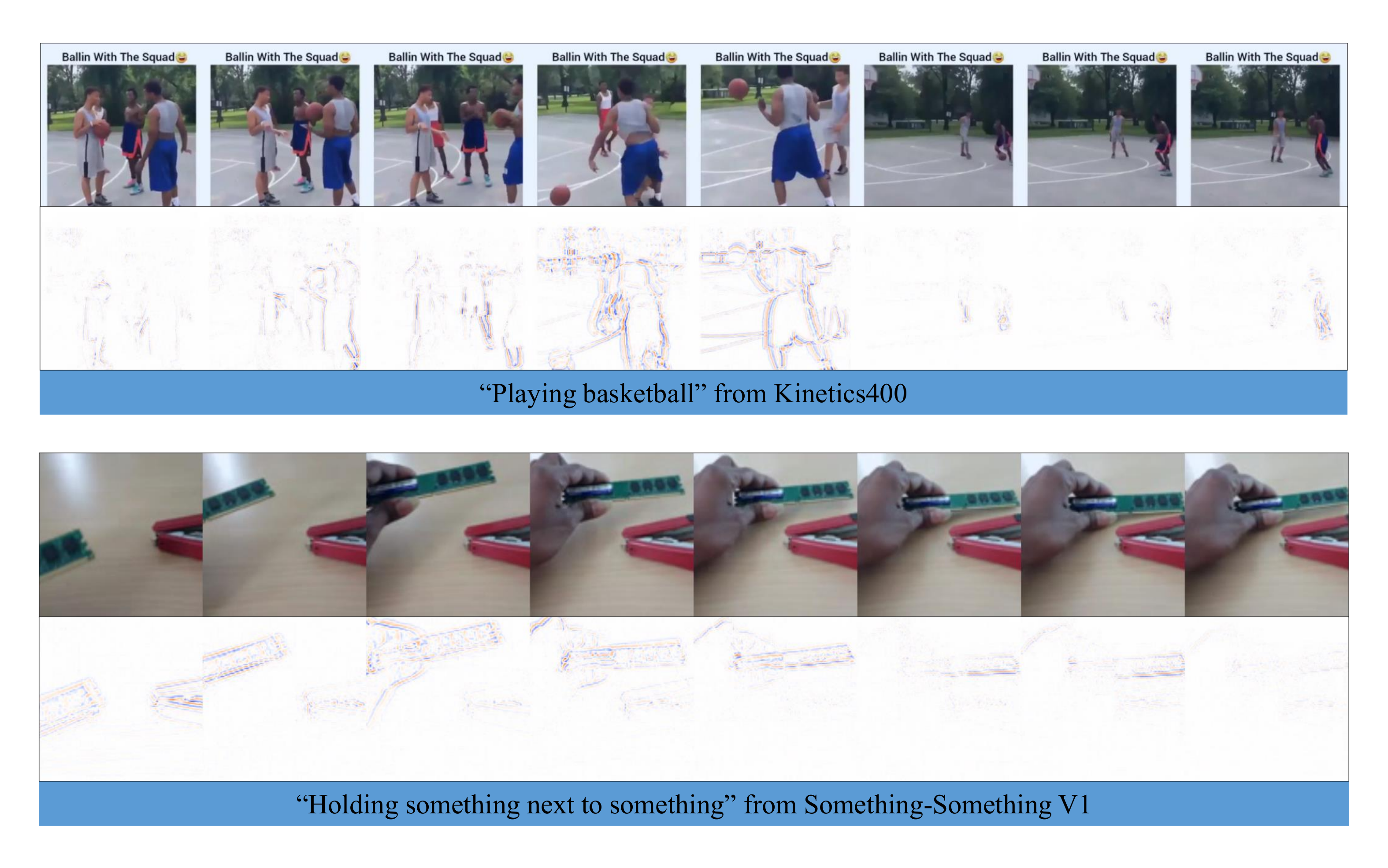}}
     \vspace*{-0.7cm}
    \caption{Examples of video and the corresponding motion representations extracted by MBPM.}
    \label{fig: more2}
\end{figure*}

These additional explanations and results are useful for Section~5.2 from the main paper.
In order to visually observe the difference between our MBPM and other motion representation method, in Figure~\ref{fig:cmpar_other_reps}, we show some example video frames and their corresponding motion representations generated by different methods. For a better view, we use the optical flow visualization approach used in~\cite{baker2011database} to visualize the output of MBPM.
The optical flow estimates the instantaneous velocity and direction of movement in every position (The color represents the direction of movement while the brightness represents the absolute value of instantaneous velocity in a position). In contrast, TVNet~\cite{fan2018end}, PA~\cite{zhang2019pan} and MBPM are more absorbed in the visual information presented in boundary regions where motion happens. 
Figures~\ref{fig: more1}-\ref{fig: more2} display the motion representation extracted by the MBPM for eight different sequences from various video datasets used for the experiments.

\section{Kernel visualization}
In Figure~\ref{fig: mbpm_kernel}, we visualize the kernel of the spatial convolution $LoG_\sigma^{1 \times k \times k}$ of MBPM in the Busy pathway. Interestingly, before and after training, kernels always present a similar shape to Mexican hats in 3-dimensional space.
In Figure~\ref{fig: visual_conv1}, we visualize the first channel of the 64 filters in the first layers of the BQN and the baseline (TSM ResNet50). We can observe that the Busy and Quiet pathways' filters have quite distinct shapes in their kernels, suggesting that the Busy and Quiet pathways learned different types of features after training.


\end{document}